\documentclass[journal]{IEEEtran}

\usepackage{blindtext}

\usepackage[table, dvipsnames]{xcolor}

\usepackage{amsmath,amssymb,amsfonts,bm}
\usepackage{lipsum}
\usepackage{times}
\usepackage{epsfig}

\usepackage{stfloats}
\usepackage{multicol}
\usepackage[inline]{enumitem}
\usepackage{algorithmic}
\usepackage{graphicx,subcaption}
\usepackage{textcomp}

\usepackage{booktabs}
\usepackage{multirow}

\usepackage[utf8]{inputenc}
\usepackage[T1]{fontenc}
\usepackage{float}
\usepackage{adjustbox}
\usepackage{array}
\usepackage{dsfont}
\usepackage{fontawesome}

\usepackage{pifont}
\newcommand{\cmark}{\ding{51}}%
\newcommand{\xmark}{\ding{55}}%
\usepackage{paralist}
\usepackage[ruled,linesnumbered]{algorithm2e}

\usepackage{arydshln}
\newcolumntype{C}[1]{>{\centering\arraybackslash}p{#1}}
\newcommand{\TF}[1]{\textcolor{blue}{#1}}
\usepackage{pifont}

\usepackage[accsupp]{axessibility}  

\usepackage{tabulary}

\definecolor{gray}{rgb}{0.3,0.3,0.3}
\definecolor{blue}{rgb}{0,0.5,1}
\definecolor{mask_red}{rgb}{1,0,0.8}
\definecolor{green}{rgb}{0.2,1,0.2}
\definecolor{rblue}{rgb}{0,0,1}

\usepackage{hyperref}
\hypersetup{colorlinks=true,linkcolor=red,citecolor=blue}

\newcommand{\revised}[1]{\textcolor{black}{#1}}

\makeatletter
\DeclareRobustCommand\onedot{\futurelet\@let@token\@onedot}
\def\@onedot{\ifx\@let@token.\else.\null\fi\xspace}

\def\eg{\emph{e.g}\onedot} 
\def\ie{\emph{i.e}\onedot}

\def\etal{\emph{et al}\onedot}

\begin{document}

\title{OAFuser: Towards Omni-Aperture Fusion for Light Field Semantic Segmentation}
\author{Fei Teng\IEEEauthorrefmark{1}, Jiaming Zhang\IEEEauthorrefmark{1}, Kunyu Peng, Yaonan Wang, Rainer Stiefelhagen, and Kailun Yang\IEEEauthorrefmark{2}%
\IEEEcompsocitemizethanks{
\IEEEcompsocthanksitem 
This work was supported in part by the National Natural Science Foundation of China (No. 62473139), in part by Hangzhou SurImage Technology Company Ltd., in part by the Ministry of Science, Research and the Arts of Baden-Württemberg (MWK) through the Cooperative Graduate School Accessibility through AI-based Assistive Technology (KATE) under Grant BW6-03, and in part by the Helmholtz Association Initiative and Networking Fund on the HAICORE@KIT and HOREKA@KIT partition.
\IEEEcompsocthanksitem F. Teng, Y. Wang, and K. Yang are with the School of Robotics and the National Engineering Research Center of Robot Visual Perception and Control Technology, Hunan University, Changsha 410082, China.
\IEEEcompsocthanksitem J. Zhang, K. Peng, and R. Stiefelhagen are with the Institute for Anthropomatics and Robotics, Karlsruhe Institute of Technology, 76131 Karlsruhe, Germany.
\IEEEcompsocthanksitem J. Zhang is also with the Institute for Visual Computing, ETH Zurich, 8092 Zurich, Switzerland.
\IEEEcompsocthanksitem \IEEEauthorrefmark{1}Equal contribution.
\IEEEcompsocthanksitem \IEEEauthorrefmark{2}Corresponding author (E-Mail: kailun.yang@hnu.edu.cn.).
}%
}

\markboth{IEEE Transactions on Artificial Intelligence, September~2024}%
{Teng \MakeLowercase{\textit{et al.}}: OAFuser}

\maketitle

%
\begin{abstract}
Light field cameras \revised{are capable of capturing} intricate angular and spatial details. This allows for acquiring complex light patterns and details from multiple angles, \revised{significantly enhancing the precision of} image semantic segmentation. However, two significant issues arise: (1) The extensive angular information of light field cameras contains a large amount of redundant data, which is overwhelming for the limited hardware resources of intelligent agents. (2) A relative displacement difference exists in the data collected by different micro-lenses. {To address these issues, we propose an \emph{Omni-Aperture Fusion model ({OAFuser})} that leverages dense context from the central view and extracts the angular information from sub-aperture images to generate semantically consistent results.} \revised{{To simultaneously streamline the redundant information} from the light field cameras and avoid feature loss during network propagation, we present a simple yet very effective \emph{Sub-Aperture Fusion Module (SAFM)}. This module efficiently embeds sub-aperture images in angular features, allowing the network to process each sub-aperture image with a minimal computational demand of only (${\sim} 1GFlops$).} {Furthermore, to address the mismatched spatial information across viewpoints, we present a \emph{Center Angular Rectification Module (CARM)} to realize feature resorting and prevent feature occlusion caused by misalignment.} {The proposed OAFuser achieves state-of-the-art performance on four UrbanLF datasets in terms of \emph{all evaluation metrics} and sets a new record of $84.93\%$ in mIoU on the UrbanLF-Real Extended dataset}, with a gain of ${+}3.69\%$. \revised{The source code for OAFuser is available at} \url{https://github.com/FeiBryantkit/OAFuser}.

\end{abstract}

\begin{IEEEImpStatement}
{\revised{To solve the data abundance problem}, we have reduced the significant computational consumption of light field cameras while not introducing any additional parameters. \revised{The proposed method} has practical value for the deployment and application of light field cameras.} The proposed method provides a scalable solution for scene parsing tasks based on light field cameras. \revised{We hope that this study will promote the development of light field cameras in the field of scene understanding.} 
\end{IEEEImpStatement}
\begin{IEEEkeywords} 
Semantic segmentation, light field, scene parsing, vision transformers, scene understanding.
\end{IEEEkeywords}

\IEEEpeerreviewmaketitle
\vspace{1em}
\section{Introduction}
\IEEEPARstart{I}{ntelligent} agents, such as robotics and autonomous driving systems, heavily rely on visual understanding, \eg, image semantic segmentation, which can produce pixel-level prediction results and contribute to determining the category, shape, and position of objects~\cite{lu2022surrogate,mazhar2023rethinking,liu2023testtime_adaptation}. To effectively apply semantic segmentation in real-world scenarios, \emph{accuracy} is one of the most crucial factors.
{To enhance the accuracy, advanced models like ConvNets~\cite{vandenhende2020mti,guo2022segnext}, MLP-based methods~\cite{tolstikhin2021mlp}, and attention mechanism~\cite{xie2021segformer,zhang2022vsa}, are introduced. Furthermore, different fusion strategies for segmentation~\cite{zhang2018exfuse,zhang2015sensor,hazirbas2017fusenet,zhang2022cmx} are proposed to improve the perception robustness.}

\begin{figure}[!t]
  \centering
  \includegraphics[width=0.48\textwidth]{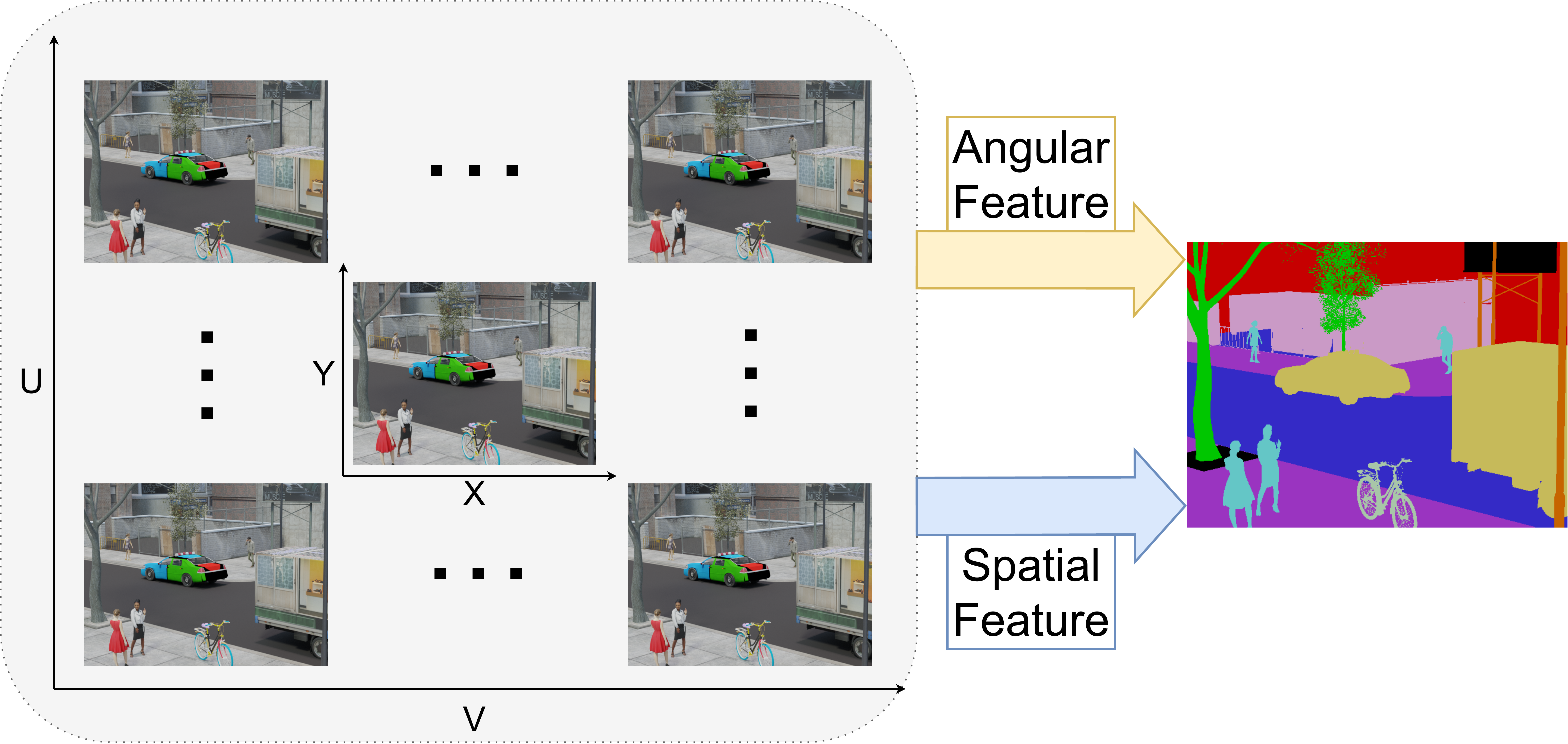}
  \caption{\revised{Comprehensive range of spatial and angular information captured using light field cameras, which contributes to pixel-wise image segmentation. The fusion of omni-aperture features improves urban semantic scene understanding.}
  }
  \label{fig:depth light}
  \vskip -2ex
\end{figure}

{In this work, we introduce Light Field (LF) cameras~\cite{georgiev2006light} for scene understanding (Fig.~\ref{fig:depth light}).}
These cameras are equipped with a unique optical system, referred to as an ``omni-aperture'', indicating their ability to utilize all available apertures to capture comprehensive light information. \revised{The volumetric radiance of a scene is recorded}, surpassing the capabilities of \revised{conventional photography} by capturing more complex data, \revised{including} information beyond the standard two-dimensional plane. {Specifically, LF cameras retain details \revised{of the angles} of incoming light rays, effectively yielding four-dimensional LF data.} By harnessing both the \textit{spatial} and \textit{angular} data, \revised{LF cameras} can facilitate \revised{postevent focus shifting} across planes \revised{thereby} enabling refined visual computations \revised{that are} critical for dense vision tasks required in vision perception tasks~\cite{meilland2015dense}.~ For intelligent agents, the ability of LF cameras to simultaneously capture spatial information from multiple directions allows for a more nuanced understanding of the environment. \revised{Furthermore, their equipment and annotation costs are lower than those of the RGB-LiDAR fusion, and LF cameras can provide more angular information than the RGB-D cameras.} 
 
Despite the theoretical benefits,
we observe two under-explored challenges \revised{when applying LF cameras to scenes semantic segmentation task}, as presented in Table~\ref{tab:2 part}, including: 
(1) {Hardware Demand and Extra Effort};
(2) The Capacity of the Network.
{To address these challenges, we propose a novel \textbf{Omni-Aperture Fusion (OAFuser)} model including two major insights for LF-based scene understanding.} 

First, hardware demand and extra effort \revised{refers} to the computational requirements necessary for processing images, \revised{which involve} the number of floating-point operations (GFlops). Given that LF cameras capture multiple SAIs in a single-shot sample, parallel processing of each image \revised{can lead} to computational overflow. This severely restricts the application of LF cameras on mobile devices.~{For instance, using an LF camera with a spatial resolution of $640{\times}480$ and angular resolution of $9{\times}9$ \revised{produces a macro-pixel image} with $4320{\times}5760$, imposing $81$ times of data stream along the network module.}
{However, compressing LF images into a depth map~\cite{chen2020bi,sheng2022urbanlf} only reduces the computational load of the semantic segmentation module. \revised{However, it requires additional auxiliary networks}, leading to a significant increase in \revised{the number of} floating-point operations.}
\revised{In addition}, the quality of the generated depth map~\cite{leistner2022towards,yan2022light} directly \revised{affects} the segmentation performance.
{To overcome these issues and explore the efficiency of using LF cameras without the need for additional computational resources or architectural modifications, we propose a simple yet effective \textbf{Sub-Aperture Fusion Module (SAFM).}~
\revised{The proposed module} minimizes computational requirements (${\sim}1~GFlops$ per image) and introduces no \revised{additional} parameters.
\revised{When analyzing image features from different viewpoints, rich information from LF images is embedded in angular and spatial feature maps}, which are then fed into transformer blocks. Thanks to the SAFM module, \revised{the proposed} OAFuser network can achieve \revised{improved} performance while avoiding excessive computational burden.}

\begin{table*}[!t]
  \centering
  \renewcommand{\arraystretch}{1.2}
  \begin{adjustbox}{width=1\textwidth}
  \begin{tabular}{c|ccc|ccc}
    \toprule[1mm]
    \multirow{2}{*}{\textbf{Method}}& \multicolumn{3}{c|}{\textbf{Hardware Demand and Extra Effort}} & \multicolumn{3}{c}{\textbf{The Capacity of the Network}} \\ \cline{2-7}
    & \multicolumn{1}{c|}{\textbf{\#Modality}} & \multicolumn{1}{c|}{{\textbf{P-SAI GFlops ($\downarrow$)}}} & \textbf{Independent of {Depth} Map} & \multicolumn{1}{c|}{\textbf{Angular Information}} & \multicolumn{1}{c|}{\textbf{Feature Rectification}} & \textbf{{Robustness} %
    } \\ \midrule[1.5pt]\hline
    RGB Based Network & 1 & N.A. & N.A. & \xmark & \xmark & \xmark \\ \hline
    RGB-D based Network & 2 & N.A. & \xmark & \xmark & \cmark & \xmark \\ \hline
    OCR-LF \cite{yuan2020object} & {>2} & 6.91 & \cmark & \cmark & \xmark & \xmark \\ \hline
    PSPNet-LF \cite{zhao2017pyramid} & {>2} & 13.01 & \cmark & \cmark & \xmark & \xmark \\ \hline
    LF-IENet \cite{cong2023combining} & {>2} & 144.95 & Explicitly & Implicit & \xmark & \xmark \\ \hline
    LF-IENet++~\cite{cong2024end} & {>2} & 159.39 & Explicitly & Implicit & \xmark & \xmark \\ \hline
    \rowcolor[gray]{.9}OAFuser (ours) & {Arbitrary} & 6.56 & \cmark & \cmark & \cmark & \cmark \\ \hline
  \end{tabular}
  \end{adjustbox}
  \caption{Challenges of LF semantic segmentation. \revised{Here}, \textit{\#Modality} indicates the number of sub-aperture images (SAI). The \textit{P-SAI GFlops} illustrates the consumption of GFlops for each SAI. \textit{N.A.} \revised{indicates} that SAIs are not used. \textit{Independent of Depth Map} indicates whether disparity information is required; the \textit{Angular Information} column explores the utilization of angular information from the SAIs; {The \textit{Robustness} indicates the issue of out-of-focused images from LF cameras. \revised{For LF-IENet series}, \textit{Depth Map} is \revised{generated explicitly}, and \textit{Angular Information} is hidden in the implicit branch (the angular feature is utilized through the depth map). Other methods perform multi-stage feature extraction for each image, which restricts the \revised{scalability of the network}. However, OAFuser extracts only spatial and angular features, \revised{which allows} it to handle arbitrary SAIs.}}
  \label{tab:2 part}
  \vskip -3ex
\end{table*}

\begin{figure}[!t]
  \centering
  \includegraphics[width=0.48\textwidth]{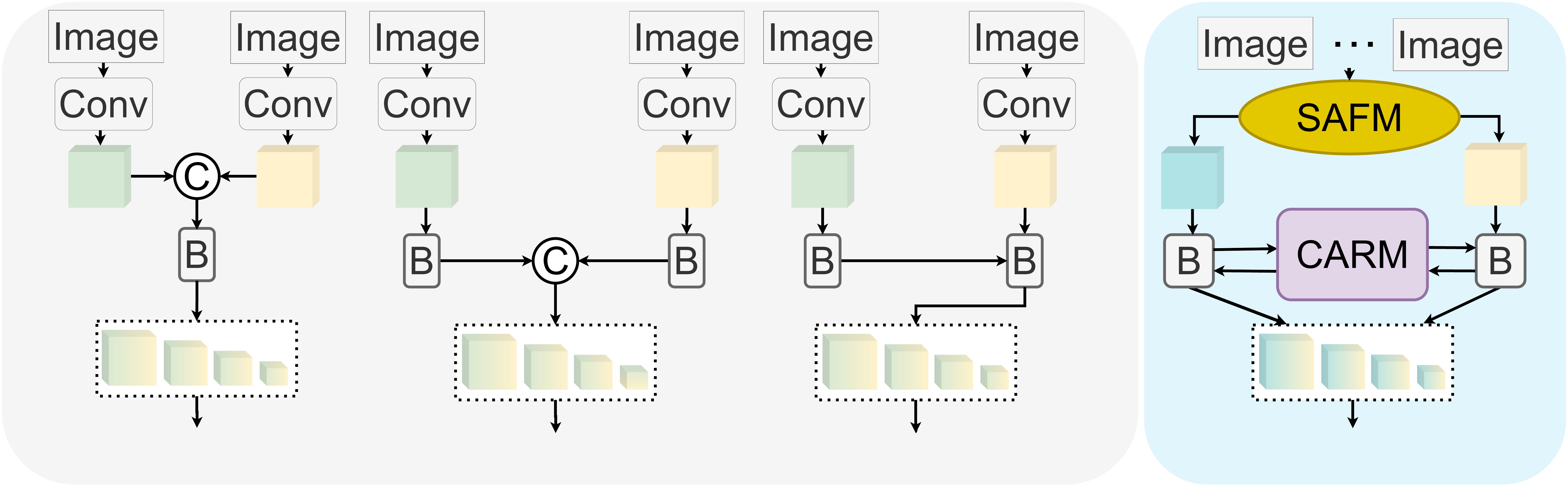}
  \begin{minipage}[t]{0.20\columnwidth}
    \vskip-3ex
    \subcaption{Early}\label{fig2_1}
  \end{minipage}%
    \begin{minipage}[t]{0.23\columnwidth}
    \vskip-3ex
    \subcaption{Late}\label{fig2_2}
  \end{minipage}%
    \begin{minipage}[t]{0.27\columnwidth}
    \vskip-3ex
    \subcaption{Injection}\label{fig2_3}
  \end{minipage}%
    \begin{minipage}[t]{0.26\columnwidth}
    \vskip-3ex
    \subcaption{Interaction}\label{fig2_4}
  \end{minipage}%
  \caption{\textbf{Paradigms of LF semantic segmentation model.} Compared \revised{with} conventional fusion methods, \revised{the} interaction fusion can handle an arbitrary number of LF images, and the features from different branches interact with each other.}
  \label{fig:fusion strategy}
  \vspace{-1em}
\end{figure}

{Second}, the capacity of the network \revised{concerns} the performance, \ie, the effective {representation} of angular features and encounter misalignment.
\revised{Previous methods that stack images into an array~\cite{sheng2022urbanlf,yan2022light} \revised{contain only} one-dimensional angular information}; the implicit angular relationship of LF images is ignored when converting them into a video sequence~\cite{zhuang2020video,wang2021temporal}. {\revised{In addition}, directly merging images or features results in blurred or mixed boundaries, and this phenomenon is particularly severe with \revised{the increasing number of sub-aperture images.}}
However, the single SAIs captured by certain micro-lenses contain \revised{a significant amount of} noise~\cite{dansereau2013decoding,bok2016geometric}. Moreover, the initial LF images are not \revised{fully} focused, which can \revised{intensify this problem}. 
In this work, we introduce a \textbf{Center Angular Rectification Module (CARM)} to perform effective rectification between the center view and the aperture-based features.
As shown in Fig.~\ref{fig:fusion strategy}, \revised{previous works} employed early-fusion strategies~\cite{10075555}, late-fusion strategies~\cite{hu2019acnet,zhang2019residual}, or injection methods~\cite{yan2022light}, \revised{which are not suitable for light field images with pixel displacement}. Our network adopts iteration feature rectification to incorporate the misaligned angular feature and spatial features from the SAFM before the fusion stage, as \revised{illustrated in} Fig.~\ref{fig2_4}.
This design allows the proposed network to explore the relationship between different features while realigning the angular and spatial information, which contributes to the performance.

{To demonstrate the effectiveness of the proposed OAFuser architecture, we conduct comprehensive experiments on different datasets. Whether on datasets with a large disparity range (UrbanLF-Syn-Big dataset) or those with a small disparity range (UrbanLF-Syn dataset) or in real-world (UrbanLF-Real dataset) or syn-real mixing (UrbanLF-RealE dataset) scenes, \revised{the proposed} method achieves state-of-the-art performance across various metrics, while using the fewest GFlops.}

At a glance, we deliver the following contributions:
\begin{compactitem}
    \item We propose a novel paradigm, \revised{the Omni-Aperture Fusion model}, \ie, \textbf{OAFuser}, to perform LF semantic segmentation by leveraging the structural characteristics of LF cameras, \revised{which addresses crucial challenges} and enables the handling of an arbitrary number of SAIs. 
    \item {We design an \textbf{Sub-Aperture Fusion Module (SAFM)} that enables our network to \revised{process each SAI using a computational cost} of (${\sim}1GFlops$) without introducing additional parameters,
    and a \textbf{Center Angular Rectification Module (CARM)} to rectify information misalignment caused by variations from different angles.}
    \item {We verify \revised{the proposed method} through extensive experiments on  four datasets, \ie,
    \revised{the} UrbanLF series.}
\end{compactitem}

\begin{figure*}[t!]
  \centering
  \includegraphics[width=1\textwidth]{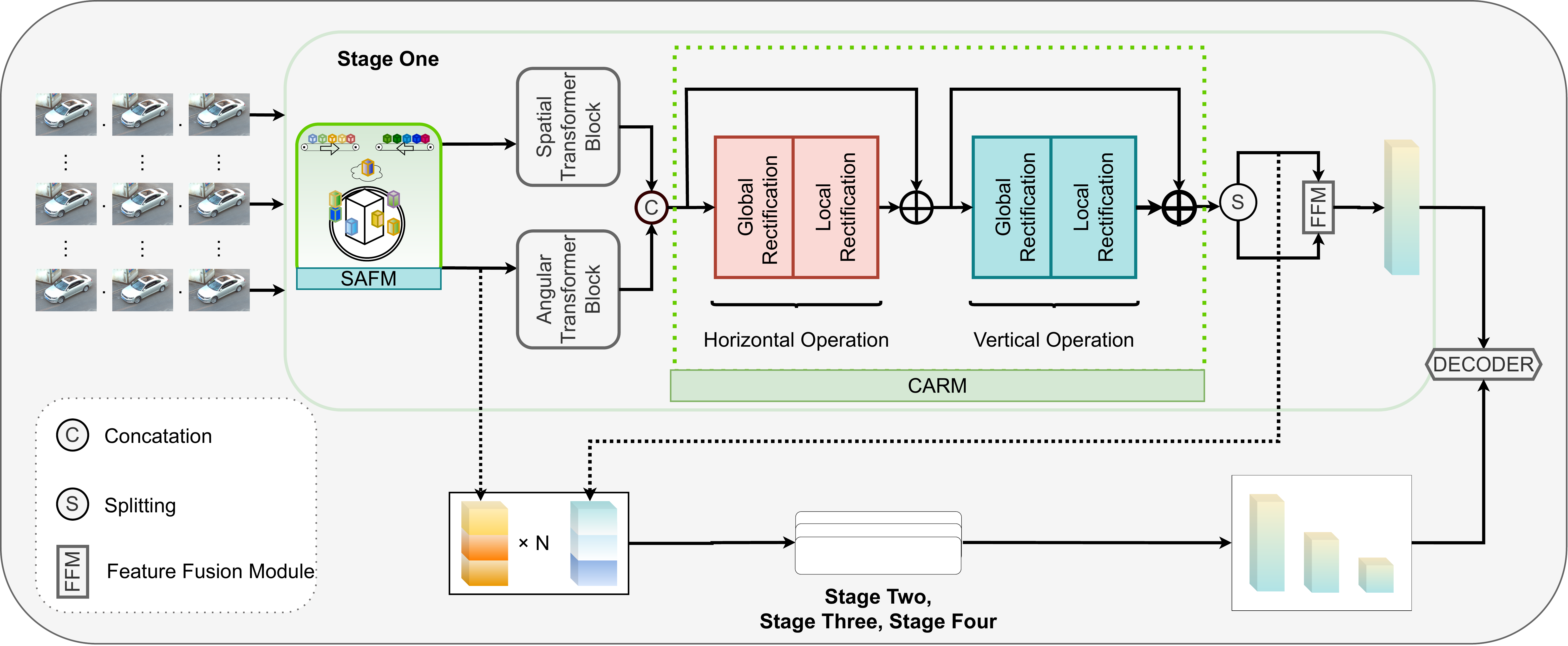}
  \caption{\revised{\textbf{Overall structure of the proposed OAFuser model.}} An arbitrary number of SAIs is input into the model. The \emph{Sub-Apeture Fusion Module (SAFM)} in Sec.~\ref{sec:3_B_sf} generates both angular and spatial features. 
  \revised{The \emph{Horizontal Operation} and \emph{Vertical Operation} of \emph{Center Aperture Rectification Module (CARM)} described in Sec.~\ref{sec:3_c_ro} are implemented for asymmetric feature rectification.} 
  Subsequently, the four-stage features are fed into the \emph{Decoder} for final semantic segmentation.
  }
  \label{fig:Overall}

\end{figure*}

\section{Related Work}
In this section, \revised{an} overview of semantic segmentation is introduced in Sec.~\ref{sec:2_a_ss} while our work is a pixel-level classification task for urban scenes. {\revised{Due to} the \revised{enormous amount} of misaligned image representation from different micro-lenses of \revised{the LF cameras}, various multi-modal semantic segmentation works are presented in Sec.~\ref{sec:2_b_mm}}. {Furthermore, to \revised{enhance the angular information} expression and better utilize the misaligned information of LF cameras, \revised{several applications of LF cameras that have significantly inspired us are also introduced}} in Sec.~\ref{sec:2_c_lf}.

\vspace{-1em}
\subsection{Semantic Segmentation}\label{sec:2_a_ss}

{Semantic scene segmentation, as a fundamental task of computer vision, plays a crucial role in scene understanding tasks, such as autonomous driving and intelligent transportation systems~\cite{zhou2022mtanet,zhang2022trans4trans}.}
Since FCN~\cite{long2015fully} pioneered the use of convolutional neural networks to replace fully connected networks with an end-to-end framework, many segmentation works have emerged based on this approach, and the efficiency and accuracy of segmentation have \revised{significantly} improved.~
For \revised{example}, the works of \cite{chen2018encoder,badrinarayanan2017segnet} adopt an encoder-decoder structure to capture contextual information and local details.
Then, dilated convolution~\cite{chen2018encoder,yang2018denseaspp} is introduced to increase the receptive field.
To enhance the global context representation, Ding \etal~\cite{ding2018context} and He \etal~\cite{he2019adaptive} adopt a pyramidal hierarchy in the encoding path.
Furthermore, the enhancement of prior contextual information~\cite{lin2017refinenet,fu2019dual,wang2021exploring} \revised{contributes to improving segmentation results.}
{Since the introduction of the self-attention mechanism in vision tasks~\cite{dosovitskiy2020image}, many subsequent works~\cite{xie2021segformer,zhang2022vsa,zhang2022trans4trans} propose dense prediction by utilizing attention-based models.
\revised{Other works introduce lightweight backbones to accelerate the inference}~\cite{sandler2018mobilenetv2,tan2019efficientnet,zhang2018shufflenet}.}

Although \revised{these studies} demonstrate excellent results in handling dense prediction tasks, they still suffer from limitations in angular information, leading to performance degradation in handling complete areas.
LF cameras, unlike monocular cameras, are equipped with an omni-aperture feature. {Therefore, LF cameras, which sample the same object from different angles, can contribute to scene understanding tasks by incorporating angular information.}

\subsection{Multi-Modal Semantic Segmentation}\label{sec:2_b_mm}
\revised{Multiple SAIs captured by an LF camera} can be considered as various RGB modalities with inherent relationships. Therefore, the research on multi-modal semantic segmentation is essential \revised{to explore} the potential of LF cameras.
ACNet~\cite{hu2019acnet} and EDCNet~\cite{zhang2021exploring} leverage attention \revised{connections to facilitate} cross-modal interactions in RGB-Depth and RGB-Event semantic segmentation, respectively.
MMFNet~\cite{chen2020mmfnet} enables the fusion of multiple medical images by aggregating different features \revised{in the spatial and channel domains.}
NestedFormer~\cite{xing2022nestedformer} \revised{proposes} a feature aggregation module to \revised{realize multi-modal medical image segmentation}.
\revised{In addition}, ESANet~\cite{seichter2021efficient} and SA-Gate~\cite{chen2020bi} utilize depth maps and RGB images to achieve high-accuracy semantic segmentation by employing fusion modules.
PGSNet~\cite{mei2022glass}, which introduces a dynamic integration module, achieves glass segmentation.
Additionally, other works~\cite{sun2019rtfnet,zhou2021gmnet,ha2017mfnet} adopt RGB-thermal image fusion.
\revised{Although CMX}~\cite{zhang2022cmx,zhang2023delivering} present an arbitrary-modal fusion network, which can handle RGB and any other modality, such as depth, thermal, polarization, event, or LiDAR data, the data sources are aligned with each other before \revised{inputting the data into the network}.~
HRFuser~\cite{broedermann2022hrfuser} realizes the fusion of an arbitrary number of additional modalities by introducing multi-window cross-attention Fig.~\ref{fig2_3}.

However, these methods are unsuitable \revised{for pixel-wise prediction of LF images}. This is because {(1)} they assign a set of computational pipelines for each modality, and \revised{applying them to LF images leads to an explosion in computational and parameter requirements}. {(2)} \revised{Pixel shifting} between different \revised{SAIs caused by micro-lenses} can adversely affect semantic segmentation results. \revised{The proposed OAFuser} focuses on reducing the computational burden of LF information and considers \revised{mismatches} between images captured by all sub-apertures of LF cameras.

\vspace{-1pt}
\subsection{Light Field Scene Understanding}\label{sec:2_c_lf}
Monocular cameras require the focal length to be set in advance during shooting. However, with LF cameras, users can adjust the focus point and refocus \revised{after capture}. \revised{Thus,} users can \revised{determine} which part of the image is sharp and which part is blurred after shooting~\cite{Ruan_2022_CVPR,yang2023joint}. \revised{Because of} their rich visual information, LF cameras have been \revised{wide applied} in various fields, such as saliency detection~\cite{wang2017two,zhang2017saliency}, depth estimation~\cite{honauer2017dataset,peng2020zero}, and super-resolution~\cite{wang2018lfnet,jin2020light}. Research in other related communities is crucial to our work, as light-field cameras are still under-explored in the semantic segmentation domain.

FES~\cite{chen2023fusion} achieves sub-aperture feature fusion via spatial and channel attention.~
NoiseLF~\cite{feng2022learning} utilizes \revised{an} {all-focus} central view image and its corresponding focal stack. It has a unique forgetting matrix and confidence re-weighting strategy to achieve supervised saliency detection \revised{in the presence of} noisy labels.~
\revised{Furthermore, several works~\cite{wang2020spatial,liang2023learning,zhang2019residual} utilize the macro-pixel image, SAIs, epipolar images, or a combination of some of these images to generate high-resolution LF images.}~In addition, AIFLFNet~\cite{zhou2023aif} utilized LF images to estimate depth information. Additionally, SAA-Net~\cite{wu2021spatial} introduces spatial-angular attention modules for LF image reconstruction.~\revised{In particular}, a design from~\cite{wang2022disentangling} proposes a unified block for handling macro-pixel images, which can be used for both super-resolution and disparity estimation.
Furthermore, \cite{9442895} has also implemented LF technology for image reconstruction in the autonomous community.

{Due to the disproportion between the computational cost and the improvements achieved, few works \revised{have focused on LF} semantic segmentation. \cite{jia2021semantic} adopts atrous spatial pyramid pooling to extract multi-scale contextual features and the angular features are acquired from the micro-pixel image. Furthermore, the UrbanLF team~\cite{sheng2022urbanlf} publishes a series of outdoor semantic segmentation datasets and utilizes \revised{stacks of SAIs to segment central-view images}. LFIE-Net~\cite{cong2023combining} introduces an explicit branch to generate disparity maps within the network and cooperates with the central view image to achieve semantic segmentation. \revised{Therefore}, LF-IENet++~\cite{10440124} is used to explore the contribution of SAIs to scene understanding in different disparity ranges.}

{However, compared with multi-modal approaches, LF-based methods, although utilizing depth maps to extract depth information, \revised{require the assistance of auxiliary networks during the depth extraction process and consume a significant amount of computational resources}, {as shown in Table~\ref{tab:Comput Cost}}. When the number of LF images increases from $5$ to $33$, the computational load of the previously best-performing method increases by more than $4000$ GFlops. However, the proposed method only adds $27.9$ Gflops of computational load.}

\section{Methodology}

This section provides a detailed introduction to \revised{the proposed network}, \ie, \emph{Omni-Aperture Fuser (OAFuser)}, which is tailored to {LF} semantic segmentation. The overall OAFuser architecture is presented in Sec.~\ref{sec:3_A_oafuser}. The \emph{Sub-Apeture Fusion Module (SAFM)} for \revised{LF} feature aggregation is introduced in Sec.~\ref{sec:3_B_sf}. The \emph{Center Aperture Rectification Module (CARM)} \revised{is presented} in Sec.~\ref{sec:3_c_ro}.  

\subsection{Proposed OAFuser Architecture}\label{sec:3_A_oafuser}

As shown in Fig.~\ref{fig:Overall}, the proposed OAFuser \revised{comprises} a four-stage encoder and a decoder. The encoder consists of the proposed SAFM and CARM to handle the 
feature fusion, feature embedding, and feature rectification, respectively.
For simplicity, the following description is based on stage one, which is \revised{the same as that for the other three stages.}
Especially, the arbitrary number of {LF} images is \revised{denoted as sub-aperture images} $F_{s_i} {\in} \mathbb{R}^{H \times W \times 3}$ and central view image $F_c{\in}\mathbb{R}^{H \times W \times 3}$, where $s_i$ \revised{denotes} the $i$-th sub-aperture image in range $[1, N]$. All \revised{images} are fed into \revised{the} SAFM to embed angular feature $F_{agl}{\in}\mathbb{R}^{\frac{H}{8} \times \frac{W}{8} \times 64}$ that contains rich angular information and spatial feature $F_{spl}{\in}\mathbb{R}^{\frac{H}{8} \times \frac{W}{8} \times 64}$ which focus on spatial information for the central view. By applying two different transformer blocks following~\cite{zhang2022cmx}, both \revised{features} are transformed into $F_{agl}^*{\in}\mathbb{R}^{\frac{H}{8} \times \frac{W}{8} \times 64}$ and $F_{spl}^*{\in}\mathbb{R}^{\frac{H}{8} \times \frac{W}{8} \times 64}$.~ {One notable aspect of our design is that the sub-aperture features are fed into subsequent stages for further processing, which \revised{differs} from other early-fusion or late-fusion methods.} {\revised{Subsequently, the angular and spatial features are concatenated and fed onto the CARM}, which includes the Horizontal Operation and the Vertical Operation for feature rectification along the horizon and vertical \revised{directions} to eliminate misalignments.} 
\revised{In particular}, the concatenation of $F_{spl}^*$ and $F_{agl}^*$ along the horizon direction is applied to obtain $F_{c1}{\in}\mathbb{R}^{\frac{H}{8} \times \frac{W}{4} \times 64}$. After the Global Rectification and the Local Rectification, the horizontally rectified feature $F^{H}{\in}\mathbb{R}^{2 \times \frac{H}{8} \times \frac{W}{8} \times 64}$ is obtained. Then, given the feature $F^{H}$, the feature $F_{c2}{\in}\mathbb{R}^{\frac{H}{4} \times \frac{W}{8} \times 64}$ is obtained by {splitting} and concatenation \revised{operations along the vertical direction}. Especially, $F^{H}{\in}\mathbb{R}^{2 \times \frac{H}{8} \times \frac{W}{8} \times 64}$ is disentangled into $F^{H}_{agl}{\in}\mathbb{R}^{\frac{H}{8} \times \frac{W}{8} \times 64}$ and $F^{H}_{spl}{\in}\mathbb{R}^{\frac{H}{8} \times \frac{W}{8} \times 64}$. $F^{H}_{agl}$ and $F^{H}_{spl}$ are further concatenated along vertical direction. After applying \revised{the Global Rectification and the Local Rectification in the Vertical Operation}, the rectified features $F^{G}{\in}\mathbb{R}^{2 \times \frac{H}{8} \times \frac{W}{8} \times 64}$. $F^{G}_{agl}{\in}\mathbb{R}^{\frac{H}{8} \times \frac{W}{8} \times 64}$ and $F^{G}_{spl}{\in}\mathbb{R}^{\frac{H}{8} \times \frac{W}{8} \times 64}$ is obtained by {split operation} of $F^{G}$. (Notably, $F^{G}_{spl}{\in}\mathbb{R}^{\frac{H}{8} \times \frac{W}{8} \times 64}$ functions \revised{serve as both a spatial feature for subsequent feature fusion and the spatial feature input for stage two.}) Furthermore, \revised{the two rectified features}, \ie,  $F^G_{agl}$ and $F^G_{spl}$ are fused by using the FFM module~\cite{zhang2022cmx}, which integrates two different features, to produce the final feature $f_1$ for this stage. The weights between \revised{the Horizontal Operation and the Vertical Operation} are shared.

Note that one of our crucial designs is \revised{that} {\textbf{all sub-aperture information is fed into each stage}, which allows our network to effectively consider all the SAIs {throughout the entire process}}. The pyramidal features $f_1, f_2, f_3, f_4$ are obtained via the four-stage encoder in a dimension of $\{64,128,320,512\}$. Subsequently, the multi-stage features are further fed into an MLP Decoder~\cite{xie2021segformer} for final prediction.

\subsection{Sub-Aperture Fusion Module}\label{sec:3_B_sf}
\begin{figure*}[t!]
  \centering
  \includegraphics[width=0.9\textwidth]{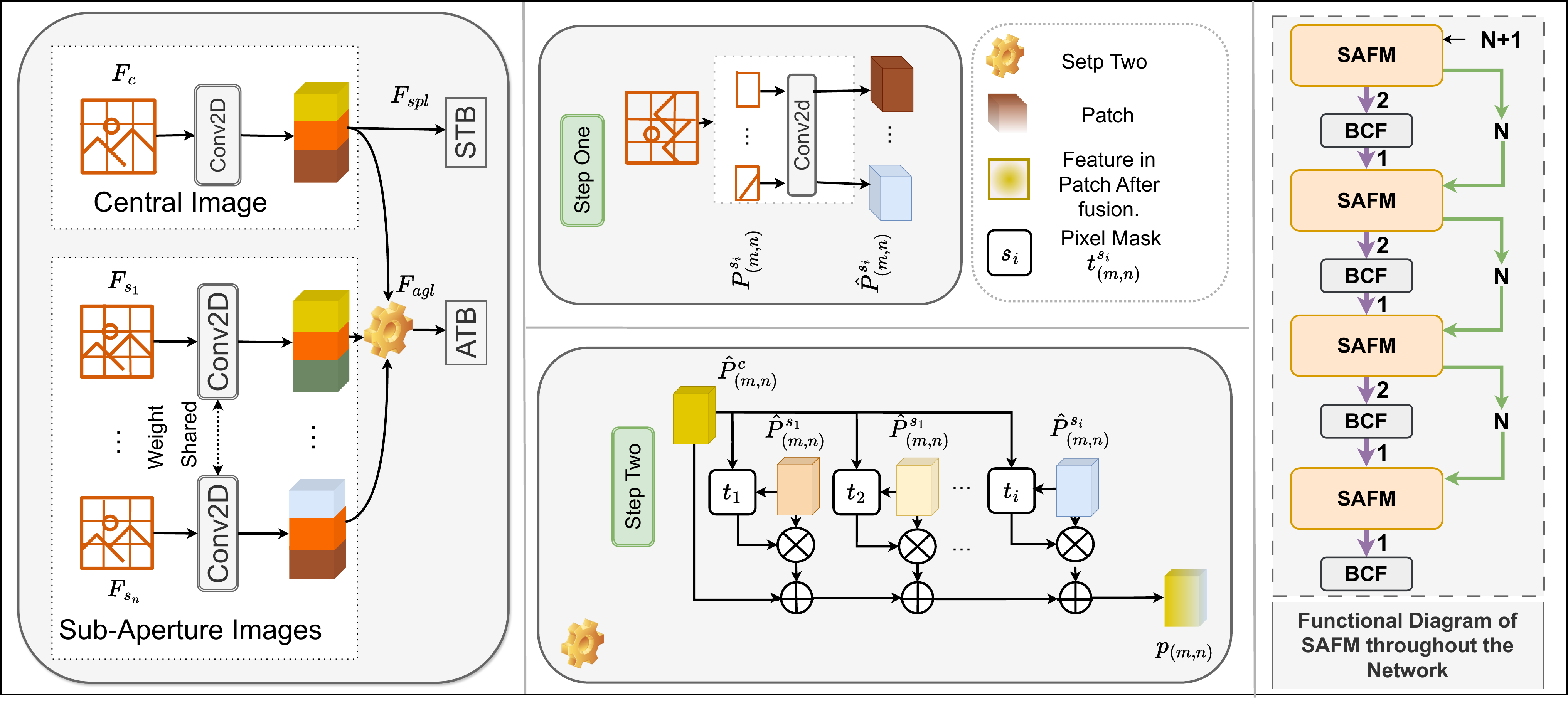}
  \caption{{\textbf{\revised{The structure of SAFM (left) and the functional diagram of the proposed SAFM (right).}} \revised{The SAFM comprises two {steps}}. Given $N+1$ images, which include $N$ $\text{SAIs}$ and one central view image, the feature aggregation is \revised{performed} in the SAFM. \revised{The SAFM} embeds the rich information into the angular feature $F_{agl}$ and the spatial feature $F_{spl}$. \emph{STB} and \emph{ATB} \revised{denote the Spatial Transformer Block and the Angular Transformer Block, respectively}. SAFM enables our network to fuse SAIs into two geometrically sensitive features, thereby avoiding the high computational pressure associated with processing \revised{each SAI individually}. \textcolor{green}{The green arrows} represent the data flow of SAIs. \textcolor{purple}{The purple arrows} indicate the paths within the backbone that require intensive calculations.}}
  \label{fig:SF Module}
\vspace{-0.5em}
\end{figure*}
To retrieve the rich angular and spatial information from {LF} images, we introduce the SAFM module, which embeds \revised{this} information into angular and spatial features. As shown in Fig.~\ref{fig:SF Module}, the SAFM \revised{comprises} of two steps. \revised{The first step extracts} features from all the {LF} images. All images are firstly divided into patches $P_{(m, n)}^{s_i}$, $P_{(m, n)}^c$, correspondence to SAIs $F_{s_i}$ and $F_c$, and further fed into the convolutional layer to extract features, resulting in $\hat{P}_{(m, n)}^{s_i}$ and $\hat{P}_{(m, n)}^c$, where $m, n$ denotes the \revised{relative patch position} in a single feature map. The patch size follows~\cite {zhang2022cmx}. Especially, this work aims to segment the central view image. Thus, the weights for the center view are \revised{calculated individually}, which ensures that the spatial information of the central view remains independent. The weights \revised{SAIs} are shared among different SAIs. The calculations are presented in Eq.~(\ref{equ:feature_extraction1}) and Eq.~(\ref{equ:feature_extraction2}):
\begin{align}
\hat{P}_{(m, n)}^{s_i} &= \text{Conv}^{sub}(C_{\text{in}}, C_{\text{out}})(P_{(m, n)}^{s_i}), \label{equ:feature_extraction1} \\ 
\hat{P}_{(m, n)}^c &= \text{Conv}^{cen}(C_{\text{in}}, C_{\text{out}})(P_{(m, n)}^c), \label{equ:feature_extraction2}
\end{align}
where $\text{Conv}(C_{in}, C_{out}) $ \revised{denotes} the convolutional layer with input dimension $C_{in}$ and output dimension $C_{out}$. For the first {step}, $C_{in}$ is $3$ and $C_{out}$ is $64$. \revised{Subsequently}, the central image features $F_{spl}$ are obtained by combining different patches $\hat{P}_{(m,n)}^c$ and fed into the Spatial Transformer Block~(STB). \revised{The patches of the central view are also fed into the next step to cooperate with patches from the SAIs to obtain the angular features.} 

Afterward, the second step of \revised{the} SAFM is illustrated in Fig.~\ref{fig:SF Module}, which \revised{is represented by} the yellow gear symbol.

{For each patch $\hat{P}_{(m, n)}^{s_i}$, the pixel score $e{_{{(m,n)}}^{s_i}}$ is calculated by obtaining the \revised{Euclidean distance} between each patch from the sub-aperture feature and the corresponding patch from the center view feature, as \revised{expressed} in Eq.~(\ref{e}):
\begin{align}
e{_{{(m,n)}}^{s_i}} &= {\text{Abs}(\hat{P}^c_{{(m, n)}}} -\hat{P}_{{(m, n)}}^{s_i}), \label{e} 
\end{align}
where $\text{Abs}{(\cdot)}$ denotes \revised{the} absolute value between two patches. Given the $e{_{{(m,n)}}^{s_i}}$, the mask score $t_{(m,n)}^{s_i}$ for $\hat{P}_{(m, n)}^{s_i}$ is obtained by mapping into the range of [1,0] and squaring them to enhance the discrimination, as shown in Eq.~(\ref{mp}).
\begin{align}
t_{{(m,n)}}^{s_i} = {(\Theta\{{e{_{{(m,n)}}^{s_i}}}}\})^2,\label{mp}
\end{align}
where $\Theta\{\cdot\}$ denotes \revised{the mapping} operation. After obtaining the mask scores, \revised{a certain patch} $P_{{(m,n)}}$ of \revised{angular features} can be calculated. \revised{Specially}, the patch $P_{{(m,n)}}$ in \revised{the angular} feature $F_{agl}$ at position $(m,n)$ can be calculated by summarizing the corresponding central view patch $\hat{P}^c_{{(m, n)}}$ with masked sub-aperture patches $\hat{P}_{{(m, n)}}^{s_i}$, as \revised{shown} in Eq.~(\ref{p_final}).
\begin{align}
P_{(m,n)} &= \hat{P}^c_{{(m, n)}} + \sum_{i=1}^{N} {t{_{{(m,n)}}^{s_i}} \cdot \hat{P}_{{(m, n)}}^{s_i}}.  \label{p_final}
\end{align}
Finally, the patch of \revised{the} angular feature $P_{(m,n)}$ is obtained. The angular feature $F_{agl}$ is filled by $P_{(m,n)}$ and further fed into \revised{the} Angular Transformer Block~(AFB).}

\subsection{Central Angular Rectification Module}\label{sec:3_c_ro}
\begin{figure}[t]
  \centering
  \includegraphics[width=0.48\textwidth]{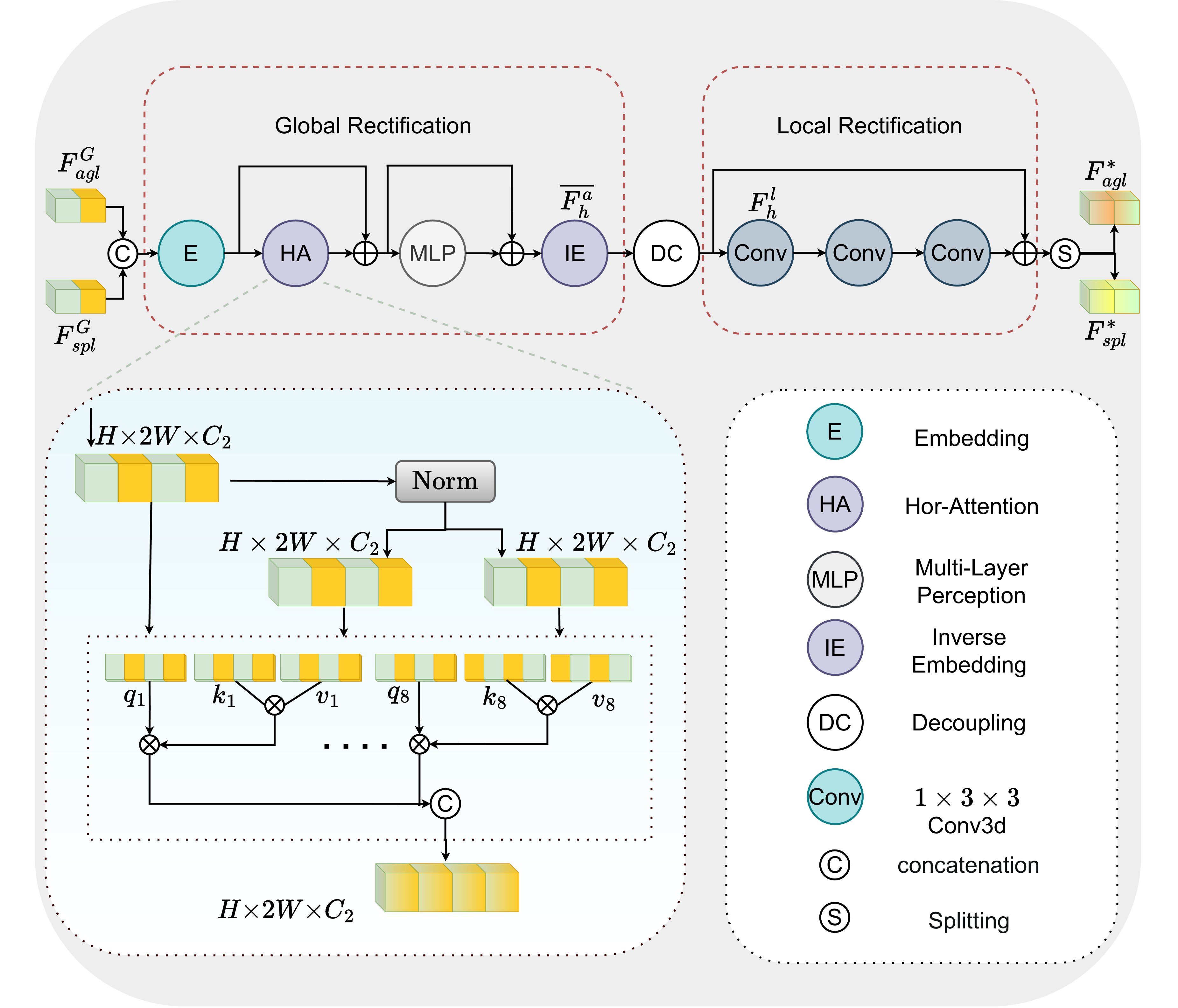}
  \caption{\textbf{The structure of \revised{the Horizontal Operation} in Center Angular Rectification Module (CARM).} It \revised{comprises} \revised{the Global Operation and the Local Operation.}}
  \label{fig:Rectification}
\vspace{-1em}
\end{figure}
{To eliminate information misalignments,} {the CARM is crucial for \revised{rectifying both spatial and angular features} while simultaneously rearranging the features captured from different viewpoints.} As shown in Fig.~\ref{fig:Rectification}, two features from different transformer blocks are fed into the CARM, which includes \revised{the} Horizontal and \revised{the} Vertical Operations. Since the Horizontal and Vertical Operations are symmetrical, only the Horizontal Operation is introduced in detail.
In each operation, \revised{the Global Rectification and the Local Rectification} are applied to rectify {feature cues} in different regions. At the end of CARM, the two features $F_{agl}^G$ and $F_{spl}^G$, which correspond to $F_{agl}^*$ and $F_{spl}^*$, are further fed into \revised{the FFM} (Feature Fusion Module). 

Specifically, in the Horizontal Operation, $F_{agl}^*$ and $F_{spl}^*$ is first concatenated along horizontal direction into feature $F_{c1} \in \mathbb{R}^{\frac{H}{8}H \times \frac{W}{4} \times 64}$. {To intuitively represent the dimension change between different steps,} we use $F_{c1} \in \mathbb{R}^{H \times 2W \times C_1}$ as the concatenated feature. \revised{First,} $F_{c1}$ is fed into the \textbf{Global Rectification} stage. Through an embedding process, $F_{c1}$ is projected \revised{onto} $F_{c1}^* \in \mathbb{R}^{H \times 2W \times C_2}$, which contain a set of tokens ${T}^*_{c1} \in \mathbb{R}^{2W \times C_2}$, with the projection matrix $M_{in} \in \mathbb{R}^{C_1 \times C_2}$, where $C_1$ denotes the input dimension of features, and $C_2$ denotes the embedding dimension for each token, and
the number of tokens is $2W$. To mitigate the covariate shift, $\dot{F}^*_{c1} \in \mathbb{R}^{H \times 2W \times C_2}$  with tokens $\dot{T}^*_{c1} \in \mathbb{R}^{2W \times C_2}$ is \revised{obtained by normalizing} $F_{c1}^*$, \revised{as described in} \cite{dosovitskiy2020image}. Then, the tokens, which represent the feature cues in a row, are utilized to generate query ($Q$), key ($K$), and value ($V$). \revised{Specifically}, $Q \in \mathbb{R}^{2W \times C_2}$ and $K \in \mathbb{R}^{2W \times C_2}$ are obtained by \revised{multiplying} tokens $\dot{T}^*_{c1}$ with matrices $M_q \in \mathbb{R}^{C_2 \times C_2}$ and $M_k \in \mathbb{R}^{C_2 \times C_2}$, respectively. The matrix $M_v \in \mathbb{R}^{C_2 \times C_2}$ multiplies the $T^*_{c1}$ to \revised{obtain} $V \in \mathbb{R}^{2W \times C_2}$. Furthermore, $Q$, $K$, and $V$ are divided along the channel dimension and fed into eight heads. The similarity of each head can be calculated by dot product and \revised{then a Softmax function} to obtain the similarity scores in each head between those tokens, \revised{as shown in} Eq.~(\ref{softmax first}).
\begin{align}
\text{Similarity}_{{head}_1} &=  \text{Softmax}\frac{Q_i\cdot K_i^T}{\sqrt{D_i} },~i. \in [1,8]. \label{softmax first}
\end{align}
Based on the similarity scores, the rectified tokens $T^a_{h_i}$ in each head can be obtained by multiplying with $V_i$, \revised{as shown} in Eq.~(\ref{softmax score}).
\begin{align}
T^a_{h_i} &= \text{Similarity}_{{head}_i} \cdot V_i,~ i \in [1,8] .\label{softmax score}
\end{align}
The final tokens $T^a_h \in \mathbb{R}^{2W \times C_2}$ are obtained by concatenation, as \revised{shown} in Eq.~(\ref{final score}).
\begin{align}
T^a_{h} &= \text{Concat}\{T^a_{h_1}, T^a_{h_2} ..., T^a_{h_8}\}.  \label{final score}
\end{align}
Given the final tokens $T_h^a$, an MLP layer is adapted, and a linear layer is to project tokens into demotion $C_1$. The final feature ${\overline{F^a_{h}}} \in \mathbb{R}^{H \times 2W \times C_1}$ after the Global Rectification can be formulated, as in the Eq.~(\ref{MLP}),
\begin{align}
    \overline{F^a_{h}} = \text{LN}(C_2,C_1)(\text{MLP}(F^a_{h}) + F^a_{h}), \label{MLP}
\end{align}
where $LN(C_2,C_1)(\cdot)$ denotes linear projection.

By decoupling the feature ${\overline{F^a_{h}}}$, the rectified feature stack $F_h^l \in \mathbb{R}^{2 \times H \times W \times C_1}$ is obtained. Furthermore, to further rectify the features in surrounding areas, the \textbf{Local Rectification} stage, which contains three convolutional layers, is implemented as in Eq.~(\ref{3d conv}), where $C$ denotes the input channel and \revised{output channels}, and $\{\cdot\}\otimes  2$ denotes repeat those layers \revised{twice}. 
\begin{align}
    \Hat{F_h^l} &= \{{LReLU}({Conv3D}(C,C)(F^l_h))\}\otimes 2,\\
    {F}^H &= {Conv3D}(C,C)(\Hat{F_h^l}).\label{3d conv}
\end{align}
By disentangling of ${F}^H\in \mathbb{R}^{2 \times H \times W \times C_1}$, $F^H_{agl} \in \mathbb{R}^{ H \times W \times C_1} $ and $F^H_{spl} \in \mathbb{R}^{H \times W \times C_1}$ is obtained and further fed into \revised{the Vertical Operation} after concatenation.

\section{Experiment Results}
\revised{To verify the effectiveness of the proposed model, we collected all UrbanLF datasets \revised{described} in Sec.~\ref{sec:4_1_da}. To ensure the reproducibility of our experiments, the training parameters are described in Sec.~\ref{sec:4_2_Im}. Quantitative and qualitative results are presented in Sec.~\ref{sec:4_3_Qua}. Note that, we achieved the best results across various datasets on various evaluation criteria.}
\subsection{Datasets}\label{sec:4_1_da}
\begin{table}[t]
\centering
\setlength{\tabcolsep}{10pt}
\renewcommand{\arraystretch}{1.5}
\begin{adjustbox}{width=0.48\textwidth}
\begin{tabular}{c|ccc|c}
\toprule[1mm]
\textbf{Dataset}           & \textbf{Train} & \textbf{Val} & \textbf{Test} & \textbf{Disparity Range} \\ \midrule[1.5pt] \hline

UrbanLF-Real      & 580   & 80  & 164  & [-0.47,1.55] \\ %
UrbanLF-Syn       & 172   & 28  & 50   & [-0.47,1.55] \\ %
UrbanLF-RealE     & 780   & 80  & 164  & [-0.47,1.55] \\ %
UrbanLF-Syn-Big   & 280   & 40  & 80   & [-7.39,7.07] \\ \hline %
\end{tabular}
\end{adjustbox}
\caption{
\textbf{Statistic information of {LF} semantic segmentation datasets.} UrbanLF-RealE denotes UrbanLF-Real with an extension of synthetic samples for training.}
\label{tab:dataset}
\end{table}

Our experiments are based on the UrbanLF datasets~\cite{sheng2022urbanlf}.
\revised{These datasets comprises} $14$ categories for urban semantic scene understanding.
Each sample in the datasets consists of $81$ SAIs with an angular resolution of $9{\times}9$.
\revised{In our experiments, we have followed the protocols proposed in UrbanLF~\cite{sheng2022urbanlf} and LFIENet++\cite{cong2024end} conducting sets of experiments: Urban-Syn, Urban-Real, UrbanLF-RealE, and UrbanLF-Syn-Big. The UrbanLF-RealE is an extension of all real and synthetic data. UrbanLF-Syn-Big has a large disparity range compared with other datasets.}
Table~\ref{tab:dataset} presents the number of images used for training, validation (quantitative and qualitative analysis), \revised{and testing} (qualitative analysis, visual comparison without ground truth), along with the disparity range.

\subsection{Implementation Details}
\begin{figure}[b]
  \centering
  \includegraphics[width=0.48\textwidth]{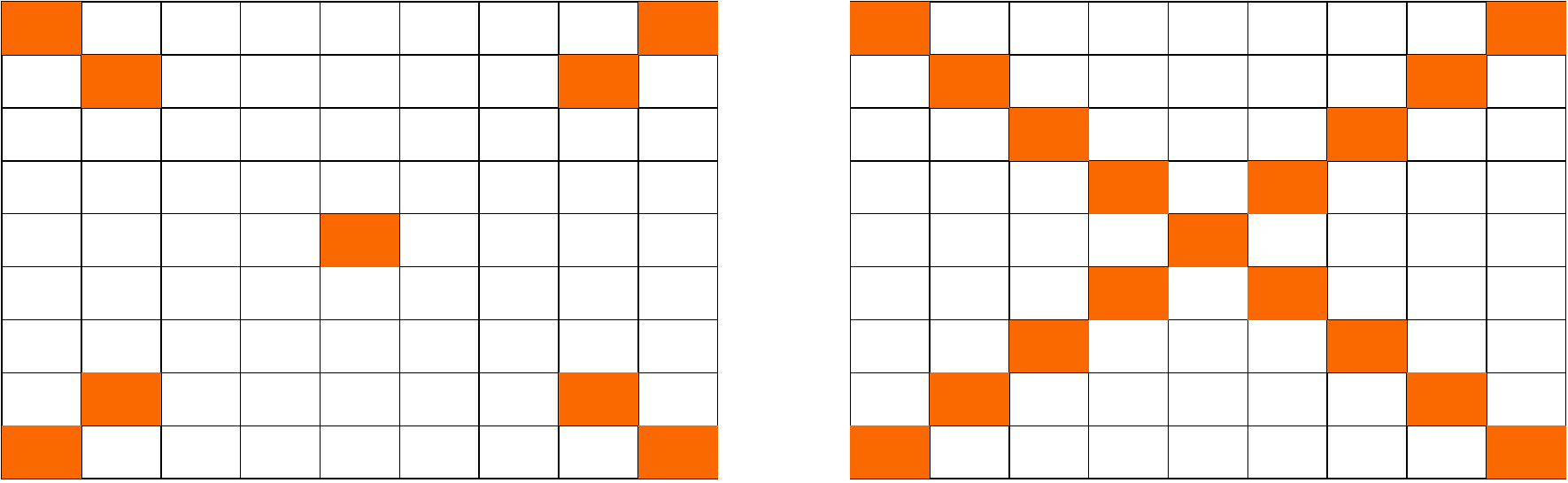}
  \begin{minipage}[t]{0.48\columnwidth}
    \vskip-3ex
    \subcaption{OAFuser9}\label{figs_a}
  \end{minipage}%
    \begin{minipage}[t]{0.48\columnwidth}
    \vskip-3ex
    \subcaption{OAFuser17}\label{figs_b}
  \end{minipage}%
  \caption{\textbf{The proposed selection strategy} of light field images, \ie, OAFuser9 (left) and OAFuser17 (right).}
  \label{fig:light array}
\end{figure}
The image size for UrbanLF-Syn is $640{\times}480$, \revised{whereas} applying zero padding converts samples in UrbanLF-Real into a size of $640{\times}480$.
Data augmentation is applied with random flipping with a probability of $0.5$, random scaling factors $\{0.5, 0.75, 1, 1.25, 1.5, 1.75\}$, normalization with mean factors $\{0.485, 0.456, 0.406\}$, and standard deviation factors $\{0.229, 0.224, 0.225\}$.
We use the AdamW optimizer with momentum parameters $\{0.9, 0.999\}$ and a weight decay of $0.01$.
The initial learning rate is set to ${6e}^{-5}$ and scheduled using the polynomial strategy with a power of $0.9$.
The first $10$ epochs are used to warm up the models.

\revised{For experiments on the four different datasets, the training process is performed on three A40 GPUs with a batch size of $3$ on each GPU, and the number of training epochs is limited to a maximum of $500$. The model is based on the MiT-B4 per-train weight~\cite{xie2021segformer}. \revised{To demonstrate model effectiveness across different} module scales, several experiments are conducted on various model scales~\ie, MiT-B4, MiT-B2, and MiT-B0.} For the ablation study of architecture, we train our model with MiT-B2~\cite{xie2021segformer} on one A5000 GPU with a batch size of $2$ and \revised{a number of} epochs of $200$ in Section~\ref{sec:5_a_model}.
With the MiT-B0~\cite{xie2021segformer} on one A5000 GPU with a batch size of $2$, we conduct an ablation study on the CARM (in Section~\ref{sec:5_a_ma}), the computational cost analysis, and the investigation on the selection of \revised{SAIs} (in Section~\ref{sec:5_d_select}).
\revised{{For the selection of SAIs, we choose images that are rich in angular information. Following the methodology of~\cite{sheng2022urbanlf,chen2023light}, the images are mainly selected along the diagonal, as shown in Fig.~\ref{fig:light array}.}}


 \label{sec:4_2_Im}

\subsection{Experiments Results} \label{sec:4_3_Qua}
To verify \revised{the proposed} methods, we compare OAFuser with other methods, including RGB-based methods~\cite{chen2018encoder,zhao2017pyramid,zheng2021rethinking,yuan2020object}, video-based light field semantic segmentation methods~\cite{zhuang2020video,wang2021temporal,jain2019accel,hu2020temporally}, and several specific designs for light field semantic segmentation~\cite{sheng2022urbanlf,cong2023combining,cong2024end} on different datasets, \ie, {UrbanLF-Real, UrbanLF-Syn, UrbanLF-Syn-Big, and UrbanLF-RealE}. \revised{Because test data are not provided, all results are based on the validation dataset. We extended the experiments of {LF-IENet} and {LF-IENet++}. OAFuser is currently the method validated on the most comprehensive dataset.}

\subsubsection{Results on UrbanLF-Real} 
\begin{table}[ht]
\centering
\renewcommand{\arraystretch}{1.4}
\setlength{\tabcolsep}{2pt}
\begin{adjustbox}{width=0.48\textwidth}
\centering
\begin{tabular}{l|c| cc| c}
\toprule[1mm]
\textbf{Method} & \textbf{Type} & \textbf{Acc (\%)} & \textbf{mAcc (\%)} & \textbf{mIoU (\%)} \\ \midrule[1.5pt]\hline

PSPNet~\cite{zhao2017pyramid} & RGB & 91.75 & 84.00 & 77.03 \\
$\text{DeepLabv3}^+$~\cite{chen2018encoder} & RGB & 91.92 & 83.45 & 76.63  \\
{SETR}~\cite{zheng2021rethinking} & RGB & 91.96 & 84.94 & 78.44\\ 
{OCR}~\cite{yuan2020object} & RGB & 92.02 & 84.99 & 78.75 \\ 
{TMANet}~\cite{wang2021temporal} & Video & 91.77 & 83.02 & 76.20 \\ 
{PSPNet-LF}~\cite{zhao2017pyramid} & LF & 91.88 & 84.31 & 77.79 \\
{OCR-LF}~\cite{yuan2020object} & LF & 92.92 & 86.50 & 80.69 \\ 
{LF-IENet$^4$}~\cite{cong2023combining} & LF & 92.40 & 84.94 & 79.03 \\ 
{LF-IENet$^3$}~\cite{cong2023combining} & LF & 93.02 & 86.69 & 80.35 \\ 
{LF-IENet++$^4$~\cite{cong2024end}}     & LF & 92.70 & 85.74 & 79.69 \\
{LF-IENet++$^3$~\cite{cong2024end}}     & LF & {\color[HTML]{0000FF}93.34} & {\color[HTML]{0000FF}87.13} & {\color[HTML]{0000FF}81.09} \\\hdashline[1pt/1pt]
\textbf{OAFuser9} & LF & {\color[HTML]{FF0000}\textbf{94.45 (+1.15)}} & {\color[HTML]{FF0000}\textbf{88.21 (+1.08)}} & {\color[HTML]{FF0000}\textbf{82.69 (+1.60)}} \\ 
\textbf{OAFuser17} & LF & 94.08 (+0.74) & 87.74 (+0.61) & 82.21 (+1.12) \\ 
\hline 
\end{tabular}
\end{adjustbox}
\caption{{Quantitative results on the UrbanLF-Real dataset. Acc (\%), mAcc (\%), and mIoU (\%) are reported. The best results are highlighted in red. The variation term \revised{represents} the performance difference from the previous best result. The superscript for ``LF-IENet'' and ``LF-IENet++'' indicates the number of sub-aperture images.}}
\label{result:real}
\end{table}

Table~\ref{result:real} presents the quantitative results \revised{obtained} on the UrbanLF-Real dataset, which is challenging due to issues \revised{such as} out-of-focus from the plenoptic camera and maintaining consistency with {LF} camera implementation without further data pre-processing in real-world scenarios.
\revised{The proposed} OAFuser9 achieves a state-of-the-art mIoU score of $82.69\%$, \revised{demonstrating} an improvement of $1.60\%$ \revised{compared with} the previous methods.
\textbf{Similarly, \revised{the proposed} OAFuser17 achieves an mIoU of $82.21\%$, with an increase of $1.12\%$. \revised{In terms of Acc and mAcc, OAFuser9 achieves an improvement of over $1\%$ compared with the previous works.}}
The small performance gap between our OAFuser9 and OAFuser17 is attributed to the image quality, which will be discussed in Sec.~\ref{sec:5_DCPC}. 

\begin{figure}[ht!]
  \centering
  \includegraphics[width=0.48\textwidth]{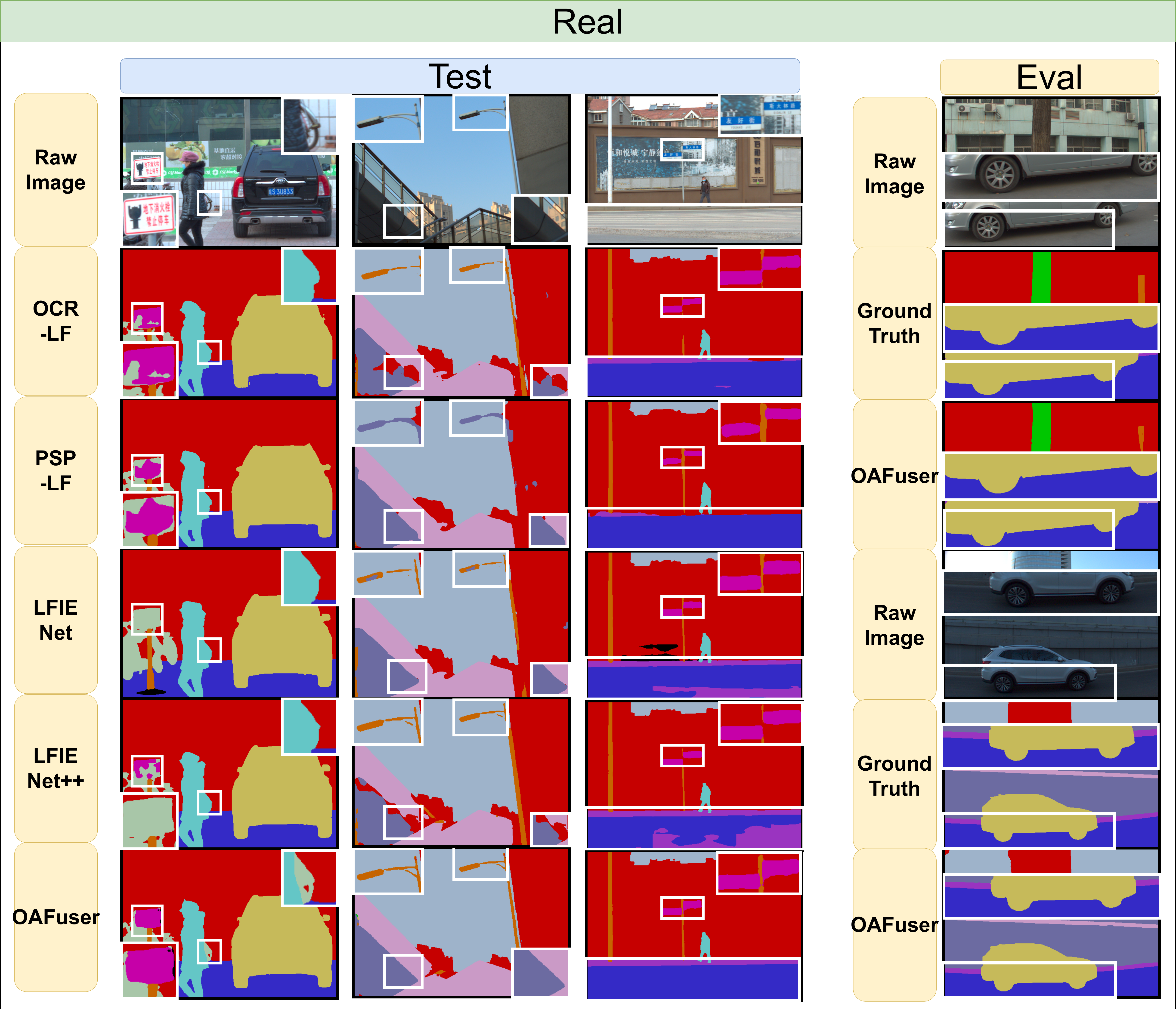}
  \caption{Qualitative \revised{result} on the UrbanLF Real dataset.}
  \label{fig:real-result}
\end{figure}

{The qualitative results, as shown in Fig.~\ref{fig:real-result}, also demonstrate the effectiveness of \revised{the proposed} method. For the three sets of images in the left column, we \revised{visually compare different methods} on the test dataset. In complex scene areas, \revised{the proposed} method achieved accurate recognition, maintaining the reasonableness of the geometric shapes of objects, \revised{whereas the previous methods} had some misjudgments. On the validation dataset, we selected challenging shadow areas. Under low-light conditions, \revised{the proposed} method also achieves accurate recognition. Furthermore, \revised{as the number of SAIs escalates, a proportional augmentation emerges in the prevalence of extraneous features within the dataset.}~{Additionally, the prevalence of out-of-focus and blurry images results in imprecise guidance, distinguishing our network from others. In \revised{the proposed method}, we meticulously implement selective embedding of features during the feature integration phase.}}
From another perspective, \revised{this} also demonstrates the necessity of the proposed method when handling semantic segmentation with {LF} cameras.
The results clearly demonstrate the remarkable effectiveness of the suggested model and support \revised{the} proposed approach, which leverages the rich angular information present in SAIs and combines it with the spatial information from the central view. This fusion of data proves to be beneficial for accurately segmenting the central view image.

\subsubsection{Results on UrbanLF-Syn} 

The quantitative results \revised{obtained} on the synthetic dataset are \revised{presented} in Table~\ref{result:syn}.  
Among all the models assessed, OAFuser17 achieves the highest mIoU. Comparatively, OAFuser9 exhibits slightly inferior performance to LF-IENet$^3$. This discrepancy in performance can be primarily attributed to the restricted size of the synthetic dataset, encompassing merely $173$ samples for training.
\begin{table}[t!]
  \centering
  \renewcommand{\arraystretch}{1.3}
  \setlength{\tabcolsep}{2pt}
  \begin{adjustbox}{width=0.48\textwidth}
\begin{tabular}{l|c|ccc}
\toprule[1mm]
\textbf{Method} & \textbf{Type} & \textbf{Acc (\%)} & \textbf{mAcc (\%)} & \textbf{mIoU (\%)} \\ \midrule[1.5pt]\hline
{PSPNet~\cite{zhao2017pyramid}} & RGB & 90.39 & 84.26 & 76.08 \\ 
$\text{DeepLabv3}^+$~\cite{chen2018encoder} & RGB & 90.89 & 84.07 & 76.45 \\ 
{OCR~\cite{yuan2020object}} & RGB & 92.37 & 87.60 & 80.13 \\ 
{MTINet~\cite{vandenhende2020mti}} & RGB-D & 92.25 & 87.46 & 80.26 \\ 
{ESANet~\cite{seichter2021efficient}} & RGB-D & 92.88 & 87.45 & 80.79 \\ 
{SA-Gate~\cite{chen2020bi}} & RGB-D & {\color[HTML]{0000FF}\text{93.03}} & 88.08 & 81.00 \\ 
{TMANet~\cite{wang2021temporal}} & Video & 90.97 & 84.72 & 76.99 \\
{PSPNet-LF~\cite{sheng2022urbanlf}} & LF & 91.16 & 86.12 & 78.48 \\ 
{OCR-LF~\cite{sheng2022urbanlf}} & LF & 92.75 & 87.98 & 80.62 \\ 
{LF-IENet$^4$~\cite{cong2023combining}} & LF & 91.39 & 86.19 & 78.50 \\ 
{LF-IENet$^3$~\cite{cong2023combining}} & LF & {92.91} & {\color[HTML]{0000FF}\text{88.21}} & {\color[HTML]{0000FF}\text{81.11}} \\ \hdashline[1pt/1pt]
\textbf{OAFuser9} & LF & 93.23 (+0.20) & \color[HTML]{FF0000}\textbf{88.26 (+0.05)} & 81.64 (+0.53) \\ 
\textbf{OAFuser17} & LF & {\color[HTML]{FF0000} \textbf{93.42 (+0.39)}} &  88.22 (+0.01) & {\color[HTML]{FF0000} \textbf{81.93 (+0.82)}} \\ \hline
\end{tabular}
\end{adjustbox}
\caption{Quantitative results on the UrbanLF-Syn dataset. Acc (\%), mAcc (\%), and mIoU (\%) are reported. The best results are highlighted in red. The variation term represents the performance difference from the previous best result.}
\label{result:syn}
\end{table}
\begin{figure}[t]
  \centering
  \includegraphics[width=0.48\textwidth]{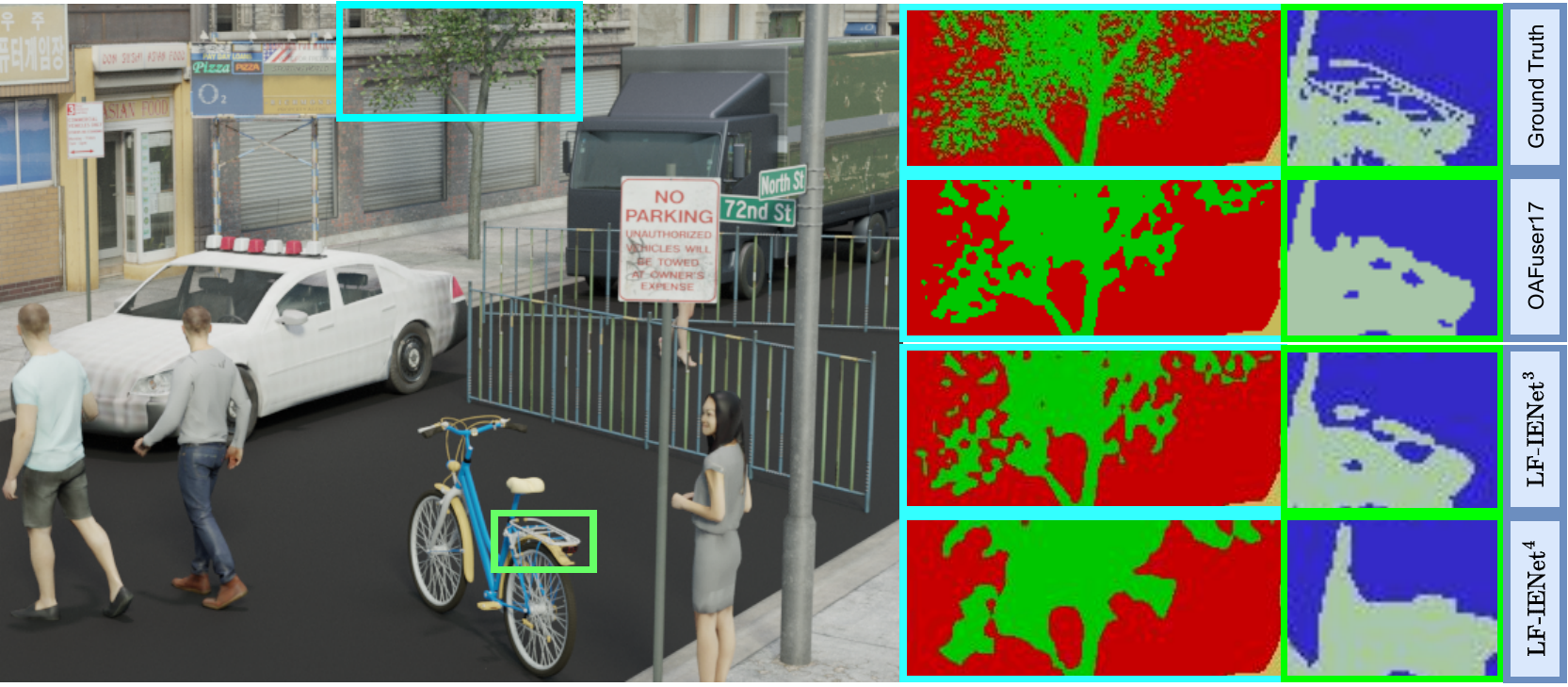}
  \caption{Qualitative \revised{results} for our network on the UrbanLF Syn dataset compared with previous methods.}
  \label{fig:Qualitative result}
\vspace{-1em}
\end{figure}

\revised{OAFuser utilizes the small differences present between different SAIs to capture intricate angular details. These details are then combined with appropriate spatial information obtained from the main viewpoint. This approach differs from previous methods that focus on spatial information limited to the central view and ignore angular information obtained from LF cameras.} {Additionally, a key feature of OAFuser is its ability to eliminate the need for pre-processing of initial {LF} images, \revised{demenstrating its ability} to handle raw {LF} data by effectively utilizing various angular information.} Importantly, \revised{the processing capabilities of the network are not adequately demonstrated when using only focused images obtained from synthetic datasets.} Furthermore, this is demonstrated in Fig.~\ref{fig:Qualitative result} in areas of complex structures, such as the leaves and bicycle pillion seats. \textbf{Compared with LF-IENet, which leads to loss of global context for objects, such as the gaps between objects and their connections with branches being disrupted, the proposed approach demonstrates better segmentation results even with $17$ SAIs.} \revised{This result highlights the superior capability of \revised{the proposed} network in harnessing SAIs}; through sophisticated rectification, the OAFuser consistently synchronizes with the rich tapestry of information gleaned from diverse perspectives. This claim is also supported by experiments conducted on other UrbanLF datasets.

\subsubsection{Results on UrbanLF-RealE} 
\begin{table}[t!]
  \centering
  \renewcommand{\arraystretch}{1.4}
  \setlength{\tabcolsep}{2pt}
  \begin{adjustbox}{width=0.48\textwidth}
\begin{tabular}{l|c|ccc}
\toprule[1mm]
\textbf{Method} & \textbf{Type} & \textbf{Acc (\%)} & \textbf{mAcc (\%)} & \textbf{mIoU (\%)} \\ \midrule[1.5pt]\hline
{PSPNet~\cite{zhao2017pyramid}} & RGB & 92.78 & 85.08 & 79.20 \\
$\text{DeepabV3}^+$~\cite{chen2018encoder} & RGB & 91.24 & 83.41 & 76.97 \\ 
{SETR~\cite{zheng2021rethinking}} & RGB & 92.69 & 86.45 & 79.87 \\ 
{OCR~\cite{yuan2020object}} & RGB & 92.93 & 86.87 & 80.43 \\ 
{TMANet~\cite{wang2021temporal}} & Video & 92.55 & 84.65 & 78.54 \\
{PSPNet-LF~\cite{zhao2017pyramid}} & LF & 92.61 & 85.22 & 79.48 \\
{OCR-LF~\cite{yuan2020object}} & LF & {\color[HTML]{0000FF}93.35} & {\color[HTML]{0000FF}87.05} & {\color[HTML]{0000FF}81.24} \\ \hdashline[1pt/1pt]
\textbf{OAFuser9} & LF & \color[HTML]{FF0000} \textbf{94.61 (+1.26)} & \color[HTML]{FF0000} \textbf{89.84 (+2.79)} & \color[HTML]{FF0000} \textbf{84.93 (+3.69)} \\ 
\textbf{OAFuser17} & LF & 93.74 (+0.39) & 88.92 (+1.87) & 82.42 (+1.18) \\ \hline
\end{tabular}
\end{adjustbox}
\caption{Quantitative results on the UrbanLF-RealE dataset. Acc (\%), mAcc (\%), and mIoU (\%) are reported. The best results are highlighted in red. The variation term indicates the performance difference from the previous best result.}
\label{result:exten}
\end{table}

UrbanLF-RealE, which involves not only real-world samples but also extends to multiple synthetic samples, poses significant challenges to the performance of the \revised{proposed} model.
This complex combination of data is more aligned with unconstrained scenarios.
As shown in Table~\ref{result:exten}, OAFuser achieves state-of-the-art performance with a mIoU of $84.93\%$.
\revised{This value exceeds those of existing methods by more than $3.69\%$.} The accuracy of OAFuser17 decreases compared \revised{with that of} OAFuser9, which is caused by the challenges posed by a large number of SAIs in complex scenarios.
\revised{With the increasing number of sub-aperture images,} it becomes challenging to accurately distinguish between relevant and irrelevant features in a dataset \revised{containing both} focused and defocused images.
Compared with PSPNet-LF and OCR-LF, which implement the direct fusion of sub-aperture and central-view images, our OAFuser embeds the rich angular feature in an independent branch and focuses on utilizing both spatial and angular features. The qualitative results are presented in Fig.~\ref{fig:RealE-Syn-result} (right).

Above all, the excellent performance \revised{obtained} on the UrbanLF-RealE dataset demonstrates the superiority of the proposed design and sets a new record of {LF} semantic segmentation.
The proposed network exhibits remarkable accuracy without relying on extensive image \revised{pre-processing} and additional devices. This validates the applicability of the proposed network in real-world scenarios.

\subsubsection{Results on UrbanLF-Syn-Big}
\begin{table}[t]
  \centering
  \renewcommand{\arraystretch}{1.4}
  \setlength{\tabcolsep}{2pt}
  \begin{adjustbox}{width=0.48\textwidth}
\begin{tabular}{l|c|ccc}
\toprule[1mm]
\textbf{Method}       & \textbf{Type}  & \textbf{Acc (\%)}   & \textbf{mAcc (\%)}  & \textbf{mIoU (\%)}  \\

\midrule[1.5pt]\hline
PSPNet~\cite{zhao2017pyramid}       & RGB   & 88.56 & 82.49 & 74.31 \\
OCR~\cite{yuan2020object}          & RGB   & 90.63 & 84.53 & 77.46 \\
SETR~\cite{zheng2021rethinking}         & RGB   & 89.37 & 84.36 & 76.29 \\
TMANet~\cite{wang2021temporal}       & Video & 88.29 & 81.13 & 73.54 \\
ESANet~\cite{seichter2021efficient}       & RGB-D & \color[HTML]{0000FF}\text{91.83} & 84.64 & 77.21 \\
SA-Gate~\cite{chen2020bi}      & RGB-D & 91.23 & 85.16 & 77.52 \\
PSPNet-LF~\cite{wang2021temporal}    & LF    & 88.78 & 83.38 & 75.15 \\
OCR-LF~\cite{sheng2022urbanlf}       & LF    & 91.09 & 84.99 & 77.90 \\
LF\_IENet$^4$~\cite{cong2023combining}   & LF    & 88.79 & 81.96 & 74.11 \\
LF\_IENet$^3$~\cite{cong2023combining}   & LF    & 91.27 & 84.71 & 78.18 \\
LF\_IENet$^4$++~\cite{cong2024end} & LF    & 89.08 & 82.55 & 74.88 \\
LF\_IENet$^3$++~\cite{cong2024end} & LF    & 91.66 & \color[HTML]{0000FF}\text{85.43} & \color[HTML]{0000FF}\text{79.21} \\ \hdashline[1pt/1pt]
\textbf{OAFuser9} & LF    & 93.39 (+1.56)    & \color[HTML]{FF0000} \textbf{89.66 (+4.23)}   & 81.15 (+1.94) \\
\text{OAFuser17}  & LF    &  \color[HTML]{FF0000} \textbf{94.29 (+2.46)}  &  88.50 (+3.07)   &   \color[HTML]{FF0000} \textbf{81.70 (+2.49)}  \\ \hline
\end{tabular}
\end{adjustbox}
\caption{Quantitative results on the UrbanLF-Syn-Big dataset. Acc ($\%$), mAcc ($\%$), and mIoU ($\%$) are reported. The best results are highlighted in red. The variation term represents the performance difference from the previous result.}%
\label{result:Syn_Big}
\vspace{-0.3em}
\end{table}
\begin{figure}[t]
  \centering
  \includegraphics[width=0.48\textwidth]{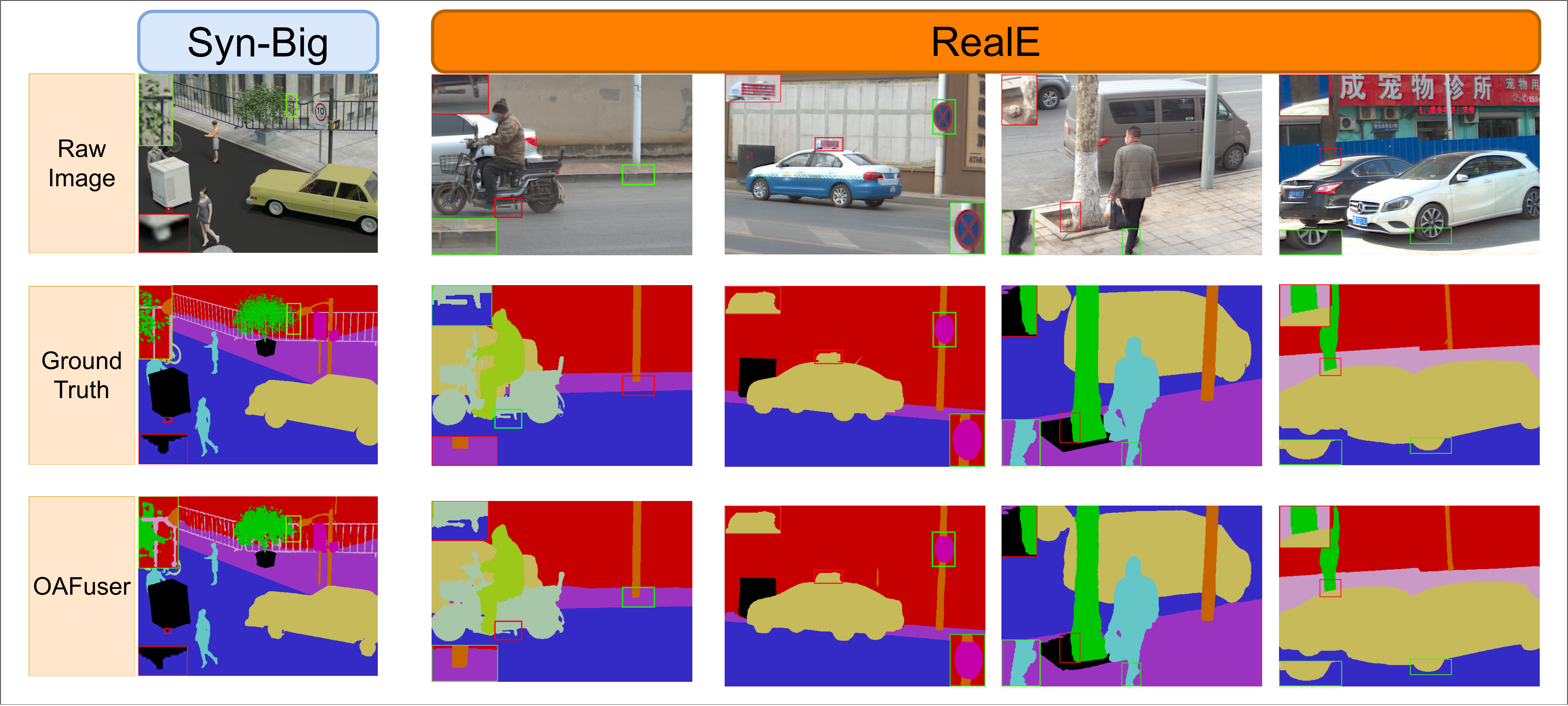}
  \caption{Qualitative result for our network on the UrbanLF RealE and Syn-Big datasets.}
  \label{fig:RealE-Syn-result}
\vspace{-0.5em}
\end{figure}
\revised{In the UrbanLF-Syn-Big dataset, the LF images exhibit an extended disparity range between adjacent views, which significantly increases the complexity of semantic segmentation tasks because of the pronounced relative pixel displacement across the image pixels.} Methods applied to this dataset have not exceeded the 80\% Acc threshold. In particular, LF-IENet and LF-IENet++ \revised{employ an auxiliary depth} estimation branch \revised{that is tailored to realize large disparity} light field semantic segmentation. However, our implementation, which integrates the SAFM and the CARM, has successfully achieved a precision of $81.70\%$. \revised{This result represents} a $2.49\%$ increment in mIoU, \revised{which is a new record for state-of-the-art performance on this challenging dataset.} By embedding high-dimensional features extracted from LF images, the utilization of angular information is ensured. Additionally, by performing feature correction in both horizontal and vertical directions, the consistency of the feature space is maintained, thereby avoiding pixel position misalignment caused by different viewpoints. As shown in Fig.~\ref{fig:RealE-Syn-result}, \revised{the corner of the table is a \revised{typically} overlooked location, and the area where plants and fences obscure each other poses a significant challenge in pixel-level classification tasks. OAFuser perfectly segmented the corner area of the table.} Although it did not perfectly reconstruct the details of the plants, it maintained the consistency of the geometric meaning between the plants and the fence.

\section{Ablation Studies}
{{In this section, \revised{we present several ablation conducted studies} to confirm the impact of different modules in the proposed method, discuss the float-point operation with the proposed network compared with other methods, and delve deeply into the network structure.}~Specifically, 
the experiments are carried out in Sec.~\ref{sec:5_a_model} to thoroughly investigate the effects of diverse components \revised{{incorporated in the proposed} method.~Furthermore, 
to assess the performance efficiency of the network}, we conduct ablation experiments, which are discussed in Sec.~\ref{sec:5_d_select}. Here, we discuss OAFuser and previous methods in terms of GFlops. \revised{In addition}, we explore the relationship between the number of SAIs and accuracy in this section.~
\revised{Furthermore, we present several experiments} in Sec.~\ref{sec:5_a_ma} to determine the optimal combination in the CARM. After that, we compare the impact of different datasets in Sec.~\ref{sec:5_DCPC} and analyze the per-class performance in Sec.~\ref{sec:5_c_per}. Moreover, a visual failure case analysis \revised{is resented} in Sec.~\ref{sec:5_e_fc}.}

\subsection{Ablation Study for the Overall Model}\label{sec:5_a_model}
\begin{table}[t]
\centering
\renewcommand{\arraystretch}{1.3}
\begin{adjustbox}{width=0.48\textwidth}
\begin{tabular}{l|cc}
\toprule[1mm]
\multicolumn{1}{c|}{\textbf{Model}} & \textbf{\#Params(M)} & \textbf{mIoU(\%)} \\ \midrule[1.5pt] \hline
\textbf{OAFuser (ours)} & 79.2 & 77.18 \\ \hdashline[1pt/1pt]
\textbf{-~Without CARM} & 65.0~(-14.2) &75.01~(-2.17) \\ \hdashline[1pt/1pt]
\textbf{-~SAIs only at Stage One} & 65.0~(-14.2) & 73.50~(-3.68) \\
\hdashline[1pt/1pt]
\textbf{-~SAIs only at Stage Four} & 65.0~(-14.2)& 73.46~(-3.72) \\
\hdashline[1pt/1pt]
\textbf{-~Baseline Method~\cite{zhang2022cmx}} & 65.0~(-14.2)& 73.25~(-3.93) \\ \hdashline[1pt/1pt]
\textbf{-~Without FFM} & 58.4~(-20.8) &70.21~(-6.93) \\ \hline
\end{tabular}
\end{adjustbox}
\caption{{Ablation study of the OAFuser framework: ``SAIs'' indicates the sub-aperture features. ``SAIs only at Stage One'' denotes that the sub-aperture features are fused in the first stage only. ``SAIs only at Stage Four'' means the SAI features are calculated but not fused and fed into the Transformer block for stages one, two, and three. The fusion (the second step \revised{in the} SAFM) occurs only in the fourth stage.}}
\label{tab:ablationstudy for model}
\end{table}
\begin{figure}[t]
  \centering
  \includegraphics[width=0.48\textwidth]{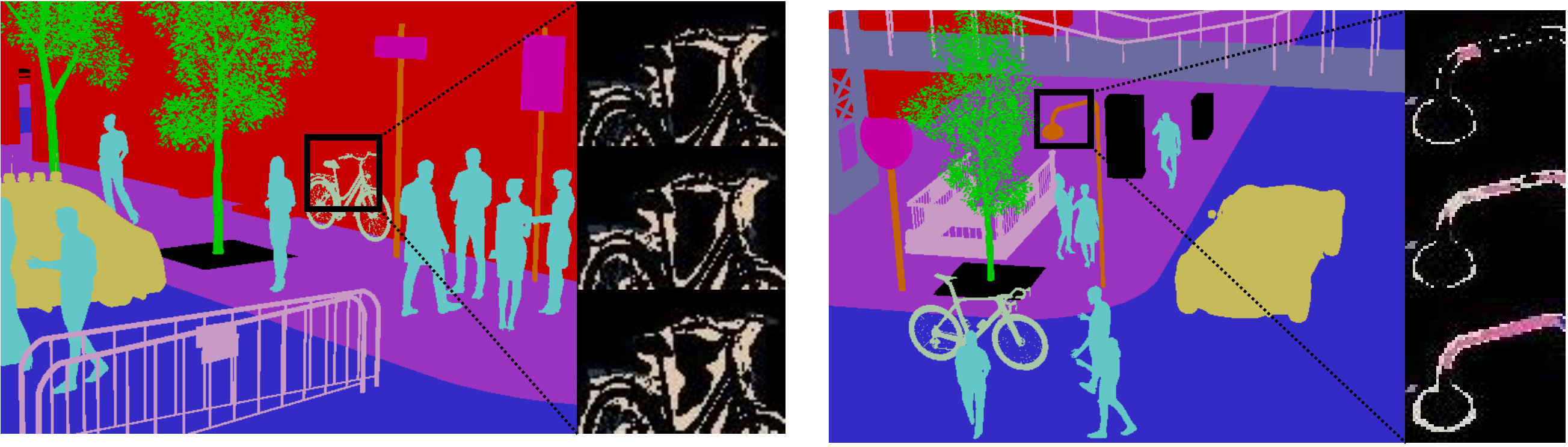}
\caption{Visualization of the proposed architecture. {From top to bottom are the entire OAFuser model, OAFuser without CARM, and OAFuser without CARM and SAFM (CMX). In the difference map, the colored area indicates the failure prediction between mask and ground truth.}}
\label{fig:ARlation}
\end{figure}

As shown in Table~\ref{tab:ablationstudy for model}, we gradually ablate the proposed OAFuser structure.
When the CARM is replaced with a feature rectification module~\cite{zhang2022cmx}, accuracy dramatically decreases by $2.17\%$.
Although the number of parameters is also reduced, the CARM module is essential to overcome challenges such as image mismatching and out-of-focus issues.~{\revised{Furthermore, we progressively ablate the second component of SAFM,} \ie, {using SAIs in the first or last stage, and without SAIs. \revised{The performance} of these three variants decreases significantly ($2{\sim}3\%$ drops \revised{compared with OAFuser}). }
Especially in these variants, the number of parameters remains unchanged, \revised{because angle information} is obtained through pixel-level feature aggregation with shared weights and \revised{addition operations.}
This also validates \revised{the claim} that the introduction of the SAFM efficiently achieves the selection and fusion of rich information from the {LF} camera, enabling the proposed network to handle arbitrary SAIs. Note that the baseline method, CMX~\cite{zhang2022cmx}, adopts a dual-pipeline SegFormer structure. Therefore, we maintain a dual-branch structure and eliminate the addition of angular features (the second step \revised{in the} SAFM) at certain stages.}
{In addition, the FFM~\cite{zhang2022cmx} is also essential because combining complementary features is crucial in the LF semantic segmentation task.}~
In addition, we conduct visualization comparisons in this ablation study. Fig.~\ref{fig:ARlation} illustrates the difference maps of the cropped region. The visual result indicates the effectiveness of using the CARM and SAFM for LF semantic segmentation. 

\subsection{Computational Cost Analysis and the Investigation \revised{of} the Selection of Sub-Aperture Images} \label{sec:5_d_select}

\begin{table}[b]
\renewcommand{\arraystretch}{1.2}
\begin{adjustbox}{width=0.48\textwidth}
\centering
\setlength{\tabcolsep}{10pt}
\begin{tabular}{c|c|c|c}
\toprule[1mm]
\textbf{Method}           & \textbf{mIoU} & \textbf{Improvement} &\textbf{P-GFlops} 
\\ \midrule[1.5pt] \hline
CMX MiT-B0~(RGB V11)~\cite{zhang2022cmx}& 68.40    & N.A. &15.49 \\  \hdashline[1pt/1pt]
OAFuser2 MiT-B0  & 71.17    & +2.77&18.71\\ 
OAFuser5 MiT-B0& 71.92   & +3.52&9.56\\
OAFuser9 MiT-B0& 72.57  & +4.17&5.90\\
OAFuser13 MiT-B0& 72.60  & +4.20&4,33 \\
OAFuser17 MiT-B0& \color[HTML]{FF0000}\textbf{72.87}  & +4.47&3.46  \\
{OAFuser21} MiT-B0& 72.22  & +3.82&3.34 \\ \hdashline[1pt/1pt]
OAFuser17~MiT-B4& \color[HTML]{FF0000} \textbf{81.93}   & +13.53&15.60 \\\hline
\end{tabular}
\end{adjustbox}
\caption{
{Exploration of the contribution of different numbers of \revised{SAIs} and the computational resource \revised{per image} (GFlops divided by \revised{the number of} SAIs). mIoU~($\%$) and P-GFlops are reported. V11 denotes the SAI from the top-left viewpoint. The best results are highlighted in red.}}
\label{tab:Sub-aperture images}
\end{table}

{One crucial aspect of \revised{the} proposed method involves utilizing each SAI to achieve better performance with a lower computational load. To verify this, we conduct intra-comparison (the comparison between OAFuser and baseline methods) and inter-comparison (the comparison between OAFuser and other methods) \revised{experiments}.}

{First, we increased the number of SAIs to record the performance and the computational cost, which is presented in Table~\ref{tab:Sub-aperture images}. Compared with the baseline method CMX~\cite{zhang2022cmx}, which is designed for \revised{similar modality fusion}, our OAFuser achieves a more than $2\%$ increase in mIoU when processing the same number of SAIs. \revised{This result} demonstrates that multi-modal fusion methods that focus on using convolution and pooling operations to process and fuse multi-modal features are not suitable for LF images with disparity differences.
Although the computational cost per image increased by $3.22$ GFlops, the average GFlops per image significantly decreases when the proposed model processes multiple images. For instance, utilizing $17$ images, the average computational demand is only $4.47$ GFlops per image, which corresponds to a $4.47\%$ improvement in accuracy. \revised{Finally, the performance of the proposed model continually improves with the incremental addition of SAIs}, further validating the efficacy of extracting and utilizing SAIs to enhance road scene understanding.}

\revised{Table.~\ref{tab:Comput Cost}}, compares the computational costs \revised{of the} proposed method and previous state-of-the-art methods.~ 
First of all, we report the GFlops consumption of OAFuser when processing different SAIs. Subsequently, we compare mIoU, Params, and GFlops based on View5 and View33 using different methods. The input image size follows the previous work~\cite{cong2024end} and is $480{\times}480$. OAFuser, when using MIT-B4 pre-trained weights to process multiple images, achieves better performance while requiring far less \revised{computations than} other methods. \revised{In particular, with $33$ images}, the computational load \revised{of the proposed model} ($216.5$ GFlops) is only $5\%$ of that of LF-IENet++ ($5259.8$ GFlops). This is attributed to the structural paradigm of OAFuser, which performs angular feature fusion \revised{to avoid} deep feature extraction for multiple images. The feature alignment through pixel-level information rectification enables the \revised{proposed} network to exhibit better performance. \revised{Although the computational load of OAFuser is slightly lower than that of OCR-LF, its performance is significantly higher.}


\begin{table}[t!]
\renewcommand{\arraystretch}{1.2}
\begin{adjustbox}{width=0.48\textwidth}
\centering

\begin{tabular}{c|c|c|c|c|c|c}
\toprule[1mm]
\#Views    & 5     & 9         & 17     & 23     & 33      & 81      \\\hline 
OAFuser    & \color[HTML]{FF0000}{\faStar}188.6 & 192.9     & 201.5  & 205.8  & 216.5   & \color[HTML]{FF0000}{\faStar}271.2   \\ \midrule[1.5pt] \hline 
\multicolumn{4}{c|}{View5}               & \multicolumn{3}{|c}{View33} \\\hline 
Method     & mIoU  & Params(M) & GFlops & mIoU   & Params  & GFlops  \\\hline 
OCR-LF     & n.a   & n.a       & n.a    & 78.26  & 137.4   & 228.1   \\
PSPNet     & n.a   & n.a       & n.a    & 75.27  & 127.8   & 429.4   \\
LF-IENet   & 79.15 & 117.4     & 719.7  & 79.23  & 117.4   & 4783.5  \\
LF-IENet++ & 79.44 & 124.7     & 920.8  & 79.59  & 124.7   & 5259.8  \\  \hdashline
OAFuser    & \TF{80.51}      & 164.1     &\color[HTML]{FF0000}{\faStar}188.6  & \textbf{80.01}      & {164.1}   & \color[HTML]{FF0000}{\faStar}216.5     \\ \hline  
\end{tabular}
\end{adjustbox}
\caption{
{\revised{Comparison of} computational consumption \revised{of the proposed} method (MIT-B4) and other methods in processing different LF images. OAFuser \revised{exhibits} a minimal increase in computational demand for the network when handling various SAIs. In scenarios with view5 and view33, \revised{owing to the} implementation of the SAFM, which prevents images from undergoing multiple feature extractions, \revised{the computational load of OAFuser} is significantly lower than that of state-of-the-art methods.}}
\vspace{-0.5em}
\label{tab:Comput Cost}
\end{table}

\subsection{Ablation Study for CARM} \label{sec:5_a_ma}

{{To rigorously assess the efficacy of \revised{the proposed} method for rectifying misaligned features, we embarked on a comprehensive suite of ablation studies centered around the innovative CARM framework.}}
\revised{As shown in Table} \ref{subtab:tableB}, we remove the local rectification module and increase the embedding dimension of the global module from $C$ to $4C$ to assess the optimal \revised{configuration}. Subsequently, upon projecting the dimensions to $2C$, our network achieves the highest score of $72.87\%$. \revised{Moreover}, excessively large dimensions (${>}2C$) might hinder the capacity to parse the features. We further explore the number of layers of 3D convolution in \revised{the Local Rectification} to evaluate \revised{its} impact.
\revised{As shown in} Table~\ref{subtab:tableA}, the \revised{use} of 3D convolutions is advantageous for feature rectification when the number of convolutional layers is ${<}4$, and the best mIoU peaks at $73.39\%$. However, when utilizing four layers of 3D convolutions, the interconnections between different features are disrupted, leading to a \revised{reduced} mIoU score of $71.49\%$.

{Subsequently, after exploring the combination of the Local Rectification Module and the Global Rectification Module through the parallel addition of features from both, \revised{we} observed a degradation in the network’s discriminative capability. This unexpected outcome also \revised{demonstrates} the effectiveness of \revised{the proposed} network.}
\begin{table}[t]
\centering

\begin{subtable}{0.48\textwidth}
\centering
\renewcommand{\arraystretch}{1.5}
\begin{tabular}{l|ccccc}
\toprule[1mm]
Dimension & C & 2C & 3C & 4C & 5C \\ \hdashline[1pt/1pt]
mIoU & 70.55 & \color[HTML]{FF0000} \textbf{72.87} & 71.52 & 71.93 & 70.57 \\ \hline
\end{tabular}
\caption{Exploration of the embedding dimension in \revised{the Global Rectification}. The \textit{Dimension} denotes \revised{the} embedding dimension.}
\label{subtab:tableB}
\end{subtable}

\hfill

\begin{subtable}{0.48\textwidth}
\centering
\renewcommand{\arraystretch}{1.5}
\begin{tabular}{l|cccccc}
\toprule[1mm] 
Layers & 0 & 1 & 2 & 3 & 4 & P \\ \hdashline[1pt/1pt]
mIoU & 72.87 & 73.13 & 73.06 & \color[HTML]{FF0000} \textbf{73.39} & 71.49 & 61.49\\ 
\hline
%
\end{tabular}
\caption{Exploration of \revised{the Local Rectification}.~\textbf{P} denotes the parallel addition of features from both groups.}
\label{subtab:tableA}
\end{subtable}

\caption{Ablation Study of the Central Angular Rectification Module~(CARM). \textit{Dimension} denotes the embedding dimension in Global Operation, and \textit{Layers} presents the number of 3D convolutions within Local Operation. mIoU (\%) is reported. The best results are highlighted in red.}
\label{tab:twosubtables}
\vspace{-0.5em}
\end{table}

\subsection{Dataset Comparison} \label{sec:5_DCPC}

\begin{table}[b]
  \centering
    \begin{adjustbox}{width=0.48\textwidth}
    \begin{tabular}{l|ccc|ccc}
    \toprule[1mm]
    Datasets & \multicolumn{3}{c|}{\#{(Real - SS)}} & \multicolumn{3}{c}{ \#(BS - SS)} \\ \midrule[1.5pt] \midrule
    \textbf{Model} & \textbf{OCR-LF} & \textbf{LF-IENet$^3$} & \textbf{OAFuser} & \textbf{OCR-LF} & \textbf{LF-IENet$^3$} & \textbf{OAFuser} \\ \hline
    
    \textbf{Acc} & -0.35  & 0.43 & \color[HTML]{FF0000} \textbf{1.22} & -1.74   & -1.64      & \color[HTML]{FF0000} \textbf{0.87}\\ 
    
    {mAcc} & -2.61  & -1.08 &  \color[HTML]{FF0000} \textbf{-0.05} & -3.07   & -3.50      & \color[HTML]{FF0000} \textbf{0.28}\\
    
    {mIoU} & -1.38 & -0.02 &  \color[HTML]{FF0000} \textbf{1.05}& -2.67   & -2.93      & \color[HTML]{FF0000} \textbf{-0.23} \\ 
    
    \bottomrule
    \end{tabular}
    \end{adjustbox}
\caption{\revised{Performance comparison} between different networks. The top three networks that perform best on the synthetic dataset are selected. \#{(Real - SS)} represents the result from the real-world dataset minus \revised{those} from the small disparity dataset, and \#{(BS - SS)} represents the result from the large disparity dataset minus \revised{those} from the small disparity dataset.~ Acc (\%), mAcc (\%), and mIoU (\%) are reported. The best results are highlighted in red.}
\label{fig:dataset comparison}
\end{table}

{To further prove the performance of \revised{the proposed network}, we analyze the impact of image quality and disparity range between distinct datasets. \revised{Although the number of classes and their distribution exhibit similarity across the datasets,} the most prominent difference lies in the image quality and disparity range, as expounded in UrbanLF~\cite{sheng2022urbanlf}.}

{We select the top three methods from the synthetic dataset for comparison (LF-IENet++ is not chosen, as it does not conduct experiments on the synthetic dataset). \textit{Image Quality:} As shown in Fig.~\ref{fig:dataset comparison}, the presence of noise in out-of-focus images \revised{results in} performance degradation for LF-IENet$^3$, and OCR-LF.
\textit{Disparity Range:} \revised{Although a significant disparity provides more information from different angles, it exacerbates misalignment between pixels, leading to poor handling of smaller categories or boundaries in images.} OCR-LF and LF-IENet \revised{exhibit} a performance decline of over $2.50\%$.
However, OAFuser \revised{exhibits} a 0.23\% decrease in mIoU, with improvements in Acc and mAcc. \revised{These improvements contribute to the high performance of the proposed method, which leverages abundant angular information and rectifies features from various viewpoints, thereby reducing the impact of image quality and disparity range on the proposed model. Regardless of noisy conditions or cases of big disparities, the proposed network exhibits superior performance.}}
\begin{table*}[ht!]
  \centering
  \LARGE
  \renewcommand{\arraystretch}{1.8}
  \begin{adjustbox}{width=1\textwidth}
\begin{tabular}{c|c|cccccccccccccccccc}
    \toprule[2mm]

\multirow{2}{*}{\textbf{Dataset}} & \multirow{2}{*}{\textbf{Method}} & \multicolumn{14}{c|}{\textbf{IoU}} & \multicolumn{1}{c|}{\multirow{2}{*}{\textbf{Acc}}} & \multicolumn{1}{c|}{\multirow{2}{*}{\textbf{mAcc}}} & \multirow{2}{*}{\textbf{mIoU}} \\ \cline{3-16}
 &  & \multicolumn{1}{c|}{\textbf{Bike}} & \multicolumn{1}{c|}{\textbf{Building}} & \multicolumn{1}{l|}{\textbf{Fence}} & \multicolumn{1}{c|}{\textbf{Others}} & \multicolumn{1}{c|}{\textbf{Person}} & \multicolumn{1}{l|}{\textbf{Pole}} & \multicolumn{1}{c|}{\textbf{Road}} & \multicolumn{1}{c|}{\textbf{Sidewalk}} & \multicolumn{1}{l|}{\textbf{Traffic Sign}} & \multicolumn{1}{c|}{\textbf{Vegetation}} & \multicolumn{1}{c|}{\textbf{Vehicle}} & \multicolumn{1}{c|}{\textbf{Bridge}} & \multicolumn{1}{c|}{\textbf{Rider}} & \multicolumn{1}{c|}{\textbf{Sky}} & \multicolumn{1}{c|}{} & \multicolumn{1}{c|}{} &  \\ \midrule[3pt]\hline
\multicolumn{1}{l|}{\multirow{3}{*}{RealE}} & \multicolumn{1}{l|}{Proportion} & 2.27 & 33.48 & 3.86 & 1.59 & 3.08 & 1.42 & 21.36 & 8.00 & 0.65 & 3.05 & 16.46 & 2.34 & 0.24 & 2.19 & n.a. & n.a. & n.a. \\ \cline{2-19}
\multicolumn{1}{l|}{} & \multicolumn{1}{l|}{CMX MiT-B4} & 86.88 & 90.52 & 76.27 & 43.47 & 91.31 & 73.47 & 90.74 & 68.70 & 86.22 & 85.21 & 96.45 & 74.33 & 69.15 & 96.61 & 93.00 & 87.00 & 80.66 \\ \cdashline{2-19}
\multicolumn{1}{l|}{} & \multicolumn{1}{l|}{OAFuser9} & 87.92 & 91.95 & 87.64 & 48.56 & 93.63 & 77.55 & 91.74 & 71.92 & 87.43 & 89.04 & 97.05 & 88.89 & 79.03 & 96.61 & 94.61 & 89.84 & \color[HTML]{FF0000} \textbf{84.93~(+4.55)} \\ \hline
\end{tabular}
\end{adjustbox}
\caption{Per-class statistics of CMX and OAFuser on the UrbanLF-RealE dataset are presented. \textit{Proportion} represents the class percentage, and the values are given in percentage (\%). The best result is highlighted in red.}
\label{tab:propotion}
\end{table*}

\subsection{Per-Class Accuracy Analysis} \label{sec:5_c_per}

\revised{To comprehensively assess the performance of the proposed model} per class and delve deeper into the improvement achieved by the model in comparison with the baseline method, we present a summary of statistical information from different methods in Table~\ref{tab:propotion}. \revised{The class proportion that assists} in data analysis is also introduced. Given that the baseline method, CMX~\cite{zhang2022cmx}, is designed for two modalities, we first aggregate the SAIs before feeding them into the network. It can be seen that OAFuser significantly outperforms the baseline method across various categories, such as \textit{fence, pole, sidewalk, vehicle, bridge} and \textit{rider}, which are crucial for the road systems in urban scenarios. \revised{Significantly, the recognition capability for \textit{vehicles} even exceeds $97\%$ in terms of IoU. } Those results \revised{demonstrate} the \revised{effectiveness} of the proposed model for segmentation tasks in urban scenes. By incorporating the SAFM and CARM, the proposed network \revised{exploits} the potential of the angle information \revised{obtained} from different SAIs and leverages the consistency of variations among SAIs. This enables OAFuser to exhibit promising performance.

\subsection{Failure Case Analysis} \label{sec:5_e_fc}
\begin{figure}[ht]
  \centering
  \includegraphics[width=0.48\textwidth]{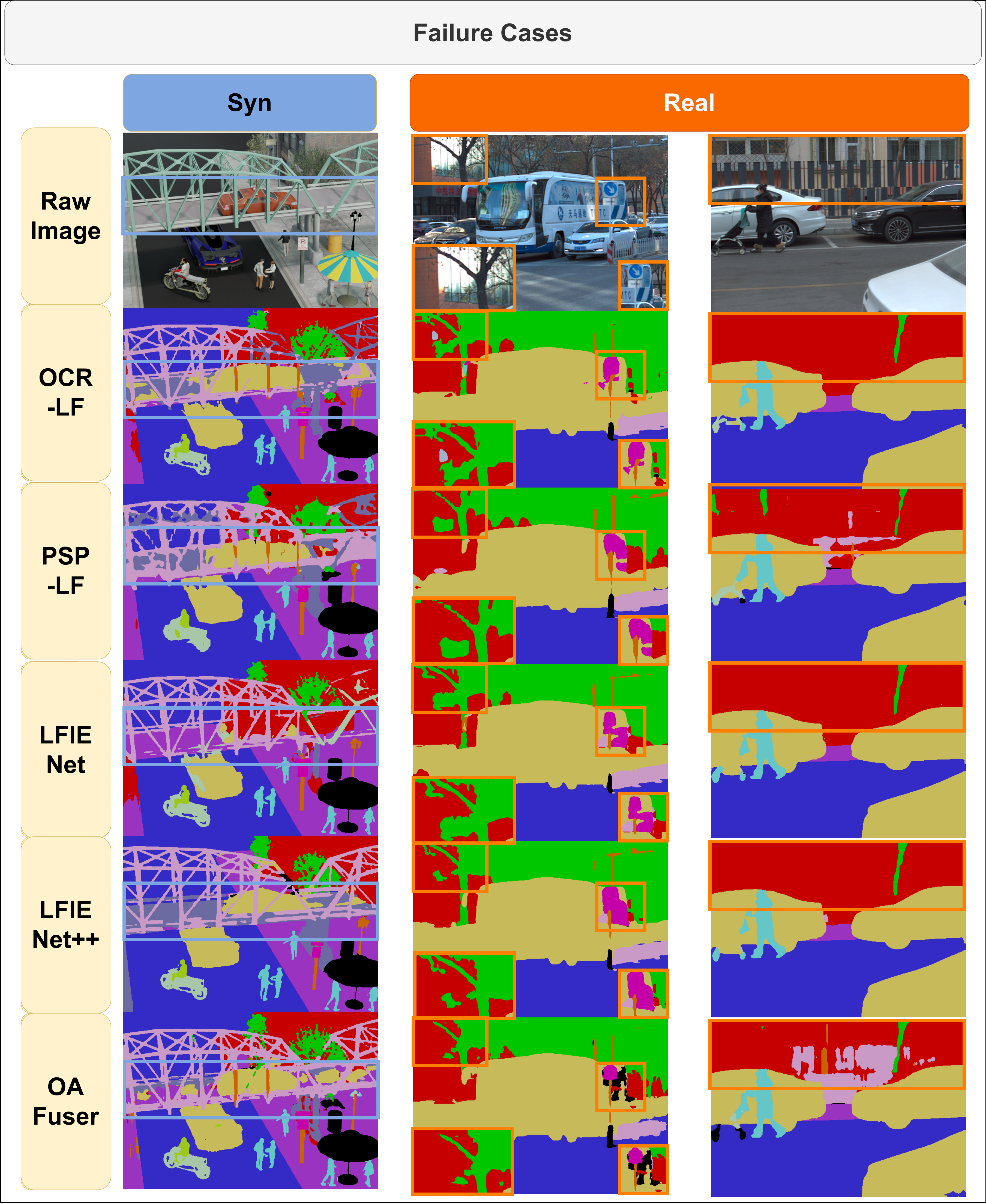}
  \caption{Failure cases across various methodologies in both synthetic big and real datasets are illustrated. Colored rectangles emphasize objects and regions that are unsegmentable.}
  \label{fig:Failure_Cases}
\vspace{-1.5em}
\end{figure}

{Although OAFuser has \revised{exhibits} state-of-the-art performance across various datasets, an analysis of the visual results presented in Fig.~\ref{fig:Failure_Cases} suggests considerable \revised{room} for improvement in complex and occluded scenarios. \revised{The results obtained} on the synthetic dataset show that all approaches demonstrate significant errors in regions where vehicles, suspension bridges, and road surfaces intersect. Although OAFuser, unlike other methods, identifies the road surface to some extent, it erroneously interprets it as a pedestrian crossing. In the middle column, where mirrors and other objects appear together in the scene, \revised{OAFuser outperforms the other methods in recognizing the signs; none of the methods recognizes the surrounding objects.} In the rightmost column, the presence of fences, due to their slender volume and frequent co-occurrence with other objects, \revised{poses a significant challenge} to the ability of the network to recognize them. \revised{The previous methods} have almost completely failed to detect fences, and OAFuser has only managed to recognize a small portion of them, which is of poor quality. In zones with \revised{overlapped vehicles, buildings, and fences}, OAFuser also exhibits recognition errors. These results demonstrate that LF semantic segmentation still requires further exploration. \revised{The proposed method significantly reduces floating-point operations and achieves state-of-the-art performance; however, further exploration is required to achieve higher accuracy and efficiency.}}

\section{Conclusion}
In this study, we explore the potential of LF cameras for road scene understanding via semantic segmentation.
We propose an innovative paradigm, the Omni-Aperture Fusion (OAFuser), \revised{to effectively exploit dense contexts} and angular information from LF apertures. 
{The Sub-Aperture Fusion Module (SAFM) is introduced, which allows the network to embed angular information from LF cameras. Each \revised{SAI incurs} minimal computational cost and does not require additional parameters.}
\revised{The introduced CARM enables the proposed network to utilize misaligned features from different viewpoints}. 
The proposed framework overcomes data redundancy limitations and establishes a new baseline for further LF exploration.

\revised{In future work, we strive to establish a benchmark with more categories and a larger set of training samples to assess the accuracy of using light field cameras for different scenarios. In particular, the implementation of LF for the indoor scenario. Moreover, how to leverage the characteristics of micro-lenses for the restoration of central images in harsh environments (\ie, heavy rain, snow) remains to be studied. Another under-explored area is exploring how to leverage numerous SAIs for other scene parsing tasks, \ie, object detection.}

\vspace{3em}
\bibliographystyle{IEEEtran}
\bibliography{bib}

\end{document}